# Stochastic Compositional Gradient Descent: Algorithms for Minimizing Compositions of Expected-Value Functions

Mengdi Wang[*]  Ethan X. Fang[*]  Han Liu[*]

**Abstract**

Classical stochastic gradient methods are well suited for minimizing expected-value objective functions. However, they do not apply to the minimization of a nonlinear function involving expected values or a composition of two expected-value functions, i.e., problems of the form $\min_x \mathbf{E}_v \left[ f_v \big( \mathbf{E}_w[g_w(x)] \big) \right]$. In order to solve this stochastic composition problem, we propose a class of stochastic compositional gradient descent (SCGD) algorithms that can be viewed as stochastic versions of quasi-gradient method. SCGD update the solutions based on noisy sample gradients of $f_v, g_w$ and use an auxiliary variable to track the unknown quantity $\mathbf{E}_w[g_w(x)]$. We prove that the SCGD converge almost surely to an optimal solution for convex optimization problems, as long as such a solution exists. The convergence involves the interplay of two iterations with different time scales. For nonsmooth convex problems, the SCGD achieve a convergence rate of $\mathcal{O}(k^{-1/4})$ in the general case and $\mathcal{O}(k^{-2/3})$ in the strongly convex case, after taking $k$ samples. For smooth convex problems, the SCGD can be accelerated to converge at a rate of $\mathcal{O}(k^{-2/7})$ in the general case and $\mathcal{O}(k^{-4/5})$ in the strongly convex case. For nonconvex problems, we prove that any limit point generated by SCGD is a stationary point, for which we also provide the convergence rate analysis. Indeed, the stochastic setting where one wants to optimize compositions of expected-value functions is very common in practice. The proposed SCGD methods find wide applications in learning, estimation, dynamic programming, etc.

## 1 Introduction

Stochastic gradient descent (SGD) methods have been prominent in minimizing convex functions by using noisy gradients. They find wide applications in simulation, distributed optimization, data-based optimization, statistical estimation, online learning, etc. SGD (also known as stochastic approximation, incremental gradient) methods have been extensively studied and well recognized as fast first-order methods which can be adapted to deal with problems involving large-scale or streaming data. For certain special cases, it has been shown that SGD exhibit the optimal sample error complexity in the statistical sense.

Classical SGD methods update iteratively by using "unbiased" samples of the iterates' gradients. In other words, the objective function is required to be linear in the sampling probabilities. Indeed, this linearity is the key to the analysis of SGD and leads to many nice properties of SGD. However, there has been little study on how to use SGD and how it performs without linearity in the sampling probabilities.

In this paper, we aim to explore the regime where *the linearity in sampling probabilities is lost*. We will develop a class of methods that we refer to as *stochastic compositional gradient methods* (SCGD), analyze their convergence properties, and demonstrate their potential applications to a broader range of stochastic problems. Consider the optimization problem

$$\min_{x \in \mathcal{X}} \Big\{ F(x) = (f \circ g)(x) \Big\}, \tag{1}$$

[*]Department of Operations Research and Financial Engineering, Princeton University, Princeton, NJ 08544, USA; e-mail: {mengdiw,xingyuan,hanliu}@princeton.edu

where $f\colon \Re^m \mapsto \Re$ is a continuous function, $g\colon \Re^n \mapsto \Re^m$ is a continuous mapping, $F\colon \Re^n \mapsto \Re$ is the function composition defined as

$$F(x) = (f \circ g)(x) = f\big(g(x)\big),$$

for all $x \in \Re^n$, and $\mathcal{X}$ is a convex and closed set in $\Re^n$. We assume throughout that there exists at least one optimal solution $x^*$ to problem (1), and we mainly focus on the case where $F = f \circ g$ is convex. However, we do no require either the outer function $f$ or the inner function $g$ to be convex or monotone. We require that $f$ be continuously differentiable, but we allow $g$ to be nonsmooth.

We are interested in situations where $f$ and $g$ take the form of expected values

$$f(y) = \mathbf{E}\left[f_v(y)\right], \qquad g(x) = \mathbf{E}\left[g_w(x)\right], \tag{2}$$

for all $y \in \Re^m, x \in \Re^n$, where $f_v\colon \Re^m \mapsto \Re$, $g_w\colon \Re^n \mapsto \Re^m$ are functions parameterized by random variables $v$ and $w$, respectively, and $\mathbf{E}\left[\cdot\right]$ is an expectation taken over the probability space of $(v, w)$. In the cases where $w$ and $v$ are independent, or more generally, where $g(x) = \mathbf{E}\left[g_w(x) \mid v\right]$ with probability 1, problem (1) becomes

$$\min_{x \in \mathcal{X}} \mathbf{E}\left[f_v\Big(\mathbf{E}\left[g_w(x) \mid v\right]\Big)\right].$$

In this paper, we consider the first-order stochastic optimization setting where the functions $f$ or $g$ cannot be directly accessed. Suppose that we have access to a Sampling Oracle ($\mathcal{SO}$) such that:

- Given some $x \in \mathcal{X}$, the oracle returns a random vector $g_w(x) \in \Re^m$.

- Given some $\bar{x} \in \mathcal{X}$, the oracle returns an $n \times m$ matrix, denoted as $\widetilde{\nabla} g_w(\bar{x}) \in \Re^{n \times m}$, whose columns are noisy gradients/subgradients vectors.

- Given some $y \in \Re^m$, the oracle returns a noisy gradient vector $\nabla f_v(y) \in \Re^m$.

More detailed specifications on the $\mathcal{SO}$ will be provided later. Our algorithmic objective is to solve problem (1)-(2) using fast updates while making queries to $\mathcal{SO}$, without storing past query responses.

The stochastic problem of the form (1)-(2) is very common in operations research applications. For one example, consider a risk averse optimization problem (see Ruszczyński and Shapiro [29] for more examples):

$$\max_x \left\{ \rho(U(x, w)) = \mathbf{E}\left[U(x, w)\right] - \mathbf{E}\Big[\left(\mathbf{E}\left[U(x, w)\right] - U(x, w)\right)_+^p\Big]^{1/p} \right\},$$

where $p > 1$ is a given scalar, in which the first term is the expected payoff and the second term is a deviation risk measure of the random payoff. When the distribution of $w$ is not known in advance, solving this problem requires Monte Carlo-based algorithms. For another example, consider the two-stage optimization problem (see the textbook by Shapiro et al. [31])

$$\min_{x_1} \mathbf{E}_{w_1}\left[\min_{x_2} \mathbf{E}_{w_2 \mid w_1}\left[U(x_1, w_1, x_2, w_2)\right]\right],$$

where $x_1$ and $x_2$ are decision variables, $w_1$ and $w_2$ are random variables which are revealed after the first and second decisions respectively, and $U$ is the utility function. This problem takes a form similar to (1)-(2), in which the inner minimization operator "$\min_{x_2}$" plays the role of the outer function $f$. The stochastic problem (1)-(2) also finds applications in statistical learning, minimax problems, dynamic programming, estimation of large deviation rate, etc (see Sections 4 and 5). Indeed, the stochastic program (1)-(2) is a ubiquitous model in practice, and is calling for efficient simulation-based algorithms.

Let us elaborate on the difficulty of problem (1)-(2) under the stochastic sampling framework. In the special case where $f$ is convex and explicitly known, one may attempt to consider the Fenchel dual representation of the problem, i.e.,

$$\min_{x \in \mathcal{X}} \sup_{\mu \in \Re^m} \left\{ \mu' \mathbf{E}[g_w(x)] - f^*(\mu) \right\},$$

where $f^*$ is the convex conjugate of $f$. However, this saddle point problem is not necessarily convex-concave (unless $g$ is affine). This excludes the possibility of applying primal-dual stochastic approximation algorithms for saddle point problems (e.g., [22]) to our composition problem (1)-(2). In the more general case where $f$ may be nonconvex or not explicitly known, the Fenchel representation is not even available.

To solve the stochastic problem (1), one may attempt to solve instead the sample average approximation (SAA) problem:

$$\min_{x \in \mathcal{X}} \frac{1}{N} \sum_{i=1}^{N} f_{v_i}\left( \frac{1}{N} \sum_{j=1}^{N} g_{w_{ij}}(x) \right),$$

in which the expectation is replaced by empirical means over some data set $\left\{ \{v_i, w_{ij}\}_{j=1}^{N} \right\}_{i=1}^{N}$. This yields a deterministic problem involving large-scale data. Two issues arise: (i) How many samples are needed for the SAA to be sufficiently accurate? (ii) How to solve the SAA problem efficiently? Regarding issue (i): As there are two nested levels of expectation in problem (1), we conjecture that the SAA might need $\mathcal{O}(N^2)$ data points to reach a certain accuracy level, which can be reached by classical SAA using $\mathcal{O}(N)$ samples if there is only one level of expectation. We will show that, by using stochastic gradient-like updates, our SCGD algorithms achieve a substantially better error sample complexity. Regarding issue (ii): The SAA problem, although deterministic, can be very difficult to solve when the data size $N^2$ is large. For big data-based problem, classical optimization techniques no longer apply as the computer memory is not sufficient to store all the data samples. To address this issue, we may interpret the SAA problem as a stochastic problem of the form (1) defined over the empirical distribution. This allows the proposed SCGD algorithms be applied to the SAA problem by processing one data point (or a small subset of data) at every iteration and making fast updates.

Now let us demonstrate why classical SGD is not applicable to problem (1)-(2). We write the gradient of $F = f \circ g$ using the chain rule (assuming that $f, g$ are differentiable):

$$\nabla F(x) = \nabla g(x) \nabla f(g(x)).$$

To apply the SGD, we need the following unbiased sample of $\nabla F$ at a given $x$, e.g.,

$$\nabla g_w(x) \nabla f_v(g(x)).$$

However, this sample gradient is not available upon one single query to $\mathcal{SO}$. This is because the value $g(x)$ is unknown for any given $x$ but can only be estimated by noisy function evaluations.

In order to address these difficulties, we are motivated to adopt a quasi-gradient approach and approximate $g(x)$ using samples. We aim to construct the approximations in such a way that they can be calculated at a minimal computational cost with small memory overhead, or even on the run. For this purpose, we propose a class of methods that update based on queries to $\mathcal{SO}$ and involve two iterations with different purposes, one for estimating $x^*$ by a stochastic gradient-like iteration and the other for maintaining a running estimate of $g(x^*)$. We refer to these methods as *stochastic compositional gradient descent* (SCGD). To guarantee the convergence of SCGD, the key observation is that we need two different stepsizes, for maintaining estimates of $x^*$ and $g(x^*)$, respectively. The estimates of $g(x^*)$ and the estimates of $x^*$ can be viewed as two entangled stochastic processes. By controlling the stepsizes, we are able to show that $\{x_k\}$ converges to some optimal solution $x^*$ with probability 1 at favorable rates.

**Related Works** The proposed SCGD is closely related to the classical stochastic gradient descent (SGD) and stochastic approximation (SA) methods. In the case where $f$ is a linear function, our method reduces to the well-known SGD or SA which has been extensively studied. Similar to several sources on SA (see e.g., the textbooks by Kushner and Yin [18], by Benveniste et al. [3], by Borkar [9], by Bertsekas and Tsitsiklis [6]), we use a supermartingale convergence argument. We use the average of iterates to improve stability and convergence rate, where the averaging idea originates from Polyak and Juditsky [25]. The idea of using multiple timescales and quasi-gradient has a long history in the literature of SA; see for example Kiefer and Wolfowitz [15], Korostelev [17], Borkar [8], Bhatnagara and Borkar [7], Konda and Tsitsiklis [16]. In the early Ukrainian literature on stochastic programming, Ermoliev [10] considered a simple two-timescale SA scheme for optimizing the composition of two expected-value functions, and showed its convergence under basic assumptions; see also [11] Section 6.7 for a brief discussion. However, there has been little further development of this idea since then, at least to the best knowledge of the authors.

The idea of stochastic gradient is also related to the class of incremental methods, which are developed for minimizing the sum of a large number of component functions. These methods update incrementally by making use of one component at a time, through a gradient-type or proximal-type iteration (see for example, [21], [20], [5]). These methods have been extended to deal with a large number of constraints $X = \cap X_i$ by random projection ([19], [34]), and also to deal with stochastic variational inequalities [33]. However, existing incremental treatments do not apply to the optimization of nonlinear functions of the sum of many component functions, which is a special case of the problem we want to address here.

Within the machine learning community, SGD for the unconstrained expectation minimization problem

$$\min_x \mathbf{E}\left[f_v(x)\right] \qquad (3)$$

has drawn significant attention. Recent advances are mostly focused on the sample error complexity of SGD in dealing with independent or adversary samples. Without smoothness assumption, it has been shown that after $k$ samples/iterations, the average of the iterates has $\mathcal{O}\left(1/k\right)$ optimization error for strongly convex objective, and $\mathcal{O}(1/\sqrt{k})$ error for general convex objective (see Rakhlin et al. [26], Shamir and Zhang [30]). To complete the picture, for nonsmooth problems with noisy gradients, there are $\Theta\left(1/k\right)$ and $\Theta(1/\sqrt{k})$ minimax information-theoretic lower bounds for convex and strongly convex problems, respectively (see e.g., Agarwal et al. [1], and see the book by Nesterov and Yudin [23] for a comprehensive study on optimization complexities). Regarding problem (3), there has been a line of works on optimal SA/SGD methods, e.g., the accelerated SA by Ghadimi and Lan [12, 13], which achieves sample error complexity that is non-improvable not only in $k$ but also in problem parameters such as the modulus of strong convexity. In contrast, to the best knowledge of the authors, there is no prior work on sample error complexity for the stochastic composition problem (1)-(2) where the objective is a nonlinear stochastic function of the expected values.

**Scope of Work** We focus on the stochastic optimization problem (1)-(2) to minimize the composition of two expected-value functions. The difficulty lies in the fact that the objective function is nonlinear in the sampling/simulation probabilities. Algorithmically, we aim to develop fast solutions that update using noisy samples. Analytically, we aim to provide theoretical guarantee in terms of optimization error and sample complexity. This study provides new insights into the level of difficulty of stochastic optimization when there exists interplay between nonlinearity and stochasticity. The main contributions of this paper are four-folded:

- We formalize the stochastic optimization problem (1)-(2), which calls for efficient algorithms that make fast updates by querying a given sampling oracle. We propose two instances of such algorithms which we refer to as stochastic compositional gradient descent (SCGD) methods. Several practical instances of problem (1)-(2) have been identified, with applications in statistical learning, dynamic programming, risk management, etc.

- The first basic SCGD algorithm that comes to our mind is related to a long-dated two-timescale SA method for the root finding problem (see Ermoliev [10]). We provide a more general proof of convergence using a coupled supermartingale convergence argument. This proof of convergence suggests that the basic SCGD applies to a broader range of problems including constrained and nonsmooth optimization. We analyze its rate of convergence in terms of optimization error after making $k$ queries to the sampling oracle, for both convex and strongly convex problems. We also prove that the limit point of SCGD must be stationary for nonconvex problems and analyze the corresponding convergence rate. To the best knowledge of the authors, this is the first result on convergence rate/sample complexity for such algorithms.

- Motivated by the basic SCGD, we propose a new accelerated algorithm which is tailored to the sampling process and has improved convergence rate for smooth optimization. The accelerated SCGD uses an additional extrapolation step before querying for a noisy function evaluation. The extrapolation step leverages the gradient continuity of smooth problems and reduces the bias in the estimates. We prove that the accelerated SCGD converges with probability 1 to an optimal solution for convex problems. More importantly, we show that the accelerated SCGD exhibits improved rate of convergence/sample convexity with the presence of smoothness in the objective function. We analyze the error sample complexity for the algorithm, in the three cases of general convex problems, strongly convex problems, and nonconvex problems, respectively.

- Our sample error complexity analysis for SCGD provides the first few benchmarks for the stochastic composition problem (1)-(2). A summary of these results is given by Table 1, and is compared with the best known results for the expectation minimization problem (3). These new benchmarks suggest that, although there are two nested levels of expectations in problem (1)-(2), we need only sub-quadratic number of samples to obtain sufficiently small optimization error.

Table 1: Summary of best known sample error complexities

|  | $\min_x (\mathbf{E}[f_v] \circ \mathbf{E}[g_w])(x)$ | | $\min_x \mathbf{E}[f_v(x)]$ |
|---|---|---|---|
|  | Non-smooth | Smooth |  |
| Convex | $\mathcal{O}(k^{-1/4})$ | $\mathcal{O}(k^{-2/7})$ | $\mathcal{O}(k^{-1/2})$ |
| Strongly Convex | $\mathcal{O}(k^{-2/3})$ | $\mathcal{O}(k^{-4/5})$ | $\mathcal{O}(k^{-1})$ |

Let us make some comparisons between analogous problems and analogous algorithms according to Table 1. First we compare the SCGD and SGD, which apply to two different problems, (1)-(2) and (3), respectively. We note that the best known convergence rate for SGD is slightly better than that of SCGD. This coincides with our intuition that the nonlinear stochastic problem (1)-(2) is generally "harder" than problem (3). Second we compare the basic SCGD and the accelerated SCGD. We note that the presence of smoothness helps the convergence rate for the stochastic method. This is consistent with the belief that nonsmooth optimization is usually "harder" than smooth optimization. An interesting insight drawn from these comparisons is: the "level of difficulty" of stochastic optimization is related to both the nonlinearity of the objective function in sampling probabilities and the smoothness of the objective in the decision variables.

The rest of the paper is organized as follows. In Section 2, we analyze the basic SCGD and prove its almost sure convergence and rate of convergence for convex, strongly convex, and nonconvex problems, respectively. In Section 3, we propose the accelerated SCGD, and we prove that it exhibits improved convergence rates for smooth optimization problems in each of the cases of convex, strongly convex, and nonconvex objectives. In Section 4, we discuss several practical instances of the stochastic program (1)-(2) to which SCGD naturally applies, including statistical learning, multi-stage optimization, estimation of rare probabilities, etc. We also discuss the extension of SCGD to the derivative-free setting where only the zeroth-order information of $f$ and $g$ is available through sampling. In Section 5, we present two numerical experiments in which we apply the SCGD methods to a statistical learning application and a dynamic programming problem, respectively.

**Notations**  All vectors are considered as column vectors. For a vector $x \in \Re^n$, we denote by $x'$ its transpose, and denote by $\|x\| = \sqrt{x'x}$ its Euclidean norm. For a matrix $A \in \Re^{n \times n}$, we denote by $\|A\| = \max\{\|Ax\| \mid \|x\| = 1\}$ its Euclidean norm. For two sequences $\{a_k\}, \{b_k\}$, we denote by $a_k = \mathcal{O}(b_k)$ if there exists $c > 0$ such that $\|a_k\| \leq \|b_k\|$ for all $k$; we denote by $a_k = \Theta(b_k)$ if there exists $c_1 > c_2 > 0$ such that $c_2\|b_k\| \leq \|a_k\| \leq c_1\|b_k\|$ for all $k$. For a set $\mathcal{X} \subset \Re^n$ and vector $y \in \Re^n$, we denote by

$$\Pi_{\mathcal{X}}\{y\} = \mathrm{argmin}_{x \in \mathcal{X}} \|y - x\|^2$$

the Euclidean projection of $y$ on $\mathcal{X}$, where the minimization is always uniquely attained if $\mathcal{X}$ is nonempty, convex and closed. For a function $f(x)$, we denote by $\nabla f(x)$ its gradient at $\mathcal{X}$ if $f$ is differentiable, and denote by $\partial f(x)$ its subdifferential at $\mathcal{X}$. We denote by "$\xrightarrow{a.s.}$" "converge almost surely to," and denote by "w.p.1" "with probability 1."

## 2 Stochastic Compositional Gradient Descent: A Basic Algorithm

We start with a basic instance of the stochastic compositional gradient descent algorithm, which we refer to as the *basic SCGD*; see Algorithm 1. This algorithm alternates between two steps: updating the solution $x_k$ by a stochastic quasi-gradient iteration, and estimating $g(x_k)$ by an iterative weighted average of past values $\{g(x_t)\}_{t=0}^{k}$.

Algorithm 1 is a most natural iteration that comes to one's mind for solving problem (1). A special case of the algorithm (when $\mathcal{X} = \Re^n$) has been considered in [10]. However, there is no known result on the convergence rate or sample complexity for this algorithm, which we establish next.

To analyze the sample complexity of Algorithm 1, we consider its averaged iterates given by

$$\widehat{x}_k = \frac{1}{N_k} \sum_{t=k-N_k}^{k} x_t,$$

**Algorithm 1** Basic SCGD
___
**Input:** $x_0 \in \Re^n$, $y_0 \in \Re^m$, stepsizes $\{\alpha_k\}, \{\beta_k\} \subset (0,1]$, $\mathcal{SO}$.
**Output:** The sequence $\{x_k\}$.
1: **for** $k = 0, 1, \ldots$ **do**
2:   Query $\mathcal{SO}$ for the sample values of $g$ at $x_k$, obtaining $g_{w_k}(x_k)$ and $\widetilde{\nabla} g_{w_k}(x_k)$.
3:   Update
$$y_{k+1} = (1 - \beta_k)y_k + \beta_k g_{w_k}(x_k).$$
4:   Query $\mathcal{SO}$ for the sample gradient of $f$ at $y_{k+1}$, obtaining $\nabla f_{v_k}(y_{k+1})$.
5:   Update
$$x_{k+1} = \Pi_{\mathcal{X}}\big\{x_k - \alpha_k \widetilde{\nabla} g_{w_k}(x_k) \nabla f_{v_k}(y_{k+1})\big\}.$$
6: **end for**
___

where $N_k \in (0, k]$ is a positive integer selected according to $k$. As long as $N_k = \Theta(k)$, the asymptotic behavior of $\widehat{x}_k$ is close to that of $x_k$ and is more stable. This is known to be true for classical stochastic approximation and stochastic gradient descent (e.g., see [25]). Throughout the rest of the paper, we take for simplicity that

$$N_k = \lceil k/2 \rceil,$$

for all $k$. For other reasonable choices of $N_k$ [e.g., $N_k = \gamma k$ for some $\gamma \in (0, 1)$], our analysis can be easily adapted to yield convergence and convergence rate results that are analogous to the current ones. A more detailed account on choices of $N_k$ and variants of averaging schemes is beyond the scope of this paper. For readers interested in this subject, we refer to the works [26, 30] and references therein.

## 2.1 Preliminaries

Throughout our discussions, we make the following assumptions.

**Assumption 1** *Let $C_g$, $C_f$, $V_g$, $L_f$ be positive scalars.*

*(i) The outer function $f$ is continuously differentiable, the inner function $g$ is continuous, and the feasible set $\mathcal{X}$ is closed and convex.*

*(ii) The random variables, $(w_0, v_0), (w_1, v_1), (w_2, v_2), \ldots$, are independent and identically distributed, such that with probability 1*

$$\mathbf{E}\left[g_{w_0}(x) \mid v_0\right] = g(x), \qquad \mathbf{E}\big[\widetilde{\nabla} g_{w_0}(x) \nabla f_{v_0}(g(x)) \mid v_0\big] \in \partial F(x), \qquad \forall\, x \in \mathcal{X}.$$

*(iii) The function $g(\cdot)$ is Lipschitz continuous with parameter $C_g$, and the samples $g_{w_k}(\cdot)$ and $\widetilde{\nabla} g_{w_k}(\cdot)$ have bounded second moments such that with probability 1*

$$\mathbf{E}\big[\|\widetilde{\nabla} g_{w_0}(x)\|^2 \mid v_0\big] \leq C_g, \quad \mathbf{E}\left[\|g_{w_0}(x) - g(x)\|^2 \mid v_0\right] \leq V_g, \quad \forall\, x \in \mathcal{X}.$$

*(iv) The functions $f$ and $f_v$ have Lipschitz continuous gradients, such that with probability 1*

$$\mathbf{E}\left[\|\nabla f_{v_0}(y)\|^2 \mid w_0\right] \leq C_f, \qquad \|\nabla f_{v_0}(y) - \nabla f_{v_0}(\bar{y})\| \leq L_f \|y - \bar{y}\|, \qquad \forall\, y, \bar{y} \in \Re^m.$$

In Algorithm 1 and Assumption 1, we intentionally do not specify what $\widetilde{\nabla} g_w(x)$ is. Let us view $\widetilde{\nabla} g_w(x)$ as a noisy sample of some generalized gradient. If $g$ is differentiable, $\widetilde{\nabla} g_w(x)$ can be a noisy gradient. If $g$ is convex but nonsmooth, $\widetilde{\nabla} g_w(x)$ can be a noisy subgradient. If $g$ is nonconvex and nonsmooth, $\widetilde{\nabla} g_w(x)$ can be some directional derivative under additional conditions. As long as the chain rule holds in expectation [Assumption 1 (ii)], i.e.,

$$\mathbf{E}\big[\widetilde{\nabla} g_{w_0}(x) \nabla f_{v_0}(g(x)) \mid v_0\big] \in \partial F(x),$$

the SCGD is guaranteed to find an optimal solution.

We denote by $\mathcal{F}_k$ the collection of random variables

$$\{x_0, \ldots, x_k, y_0, \ldots, y_k, w_0, \ldots, w_{k-1}, v_0, \ldots, v_{k-1}\}.$$

Let us first derive some basic inequalities based on Assumption 1. By using Assumption 1 (iii),(iv), we can show that $\|x_{k+1} - x_k\|^2$ is of order $\alpha_k^2$, i.e.,

$$\begin{aligned}
\mathbf{E}\left[\|x_{k+1} - x_k\|^2 \mid \mathcal{F}_k\right] &\leq \alpha_k^2 \mathbf{E}\big[\|\widetilde{\nabla} g_{w_k}(x_k)\|^2 \|\nabla f_{v_k}(y_{k+1})\|^2 \mid \mathcal{F}_k\big] \\
&= \alpha_k^2 \mathbf{E}\big[\mathbf{E}[\|\widetilde{\nabla} g_{w_k}(x_k)\|^2 \mid v_k] \|\nabla f_{v_k}(y_{k+1})\|^2 \mid \mathcal{F}_k\big] \\
&\leq \alpha_k^2 C_g C_f,
\end{aligned} \qquad (4)$$

for all $k$, with probability 1.

While analyzing Algorithm 1, we observe that the two error sequences $\{x_k - x^*\}$ and $\{g(x_k) - y_{k+1}\}$ are coupled together in their asymptotic behaviors. Then we use the following coupled supermartingale convergence lemma by Wang and Bertsekas [34] to prove their almost sure convergence. This lemma generalizes the earlier supermartingale convergence lemma by Robbins and Siegmund [28]. It characterizes the inherent convergence of two random processes that are coupled together.

**Lemma 1 (Coupled Supermartingale Convergence Lemma)** *Let $\{\xi_k\}$, $\{\zeta_k\}$, $\{u_k\}$, $\{\bar{u}_k\}$, $\{\eta_k\}$, $\{\theta_k\}$, $\{\epsilon_k\}$, $\{\mu_k\}$, and $\{\nu_k\}$ be sequences of nonnegative random variables such that*

$$\mathbf{E}\left[\xi_{k+1} \mid \mathcal{G}_k\right] \leq (1 + \eta_k)\xi_k - u_k + c\theta_k \zeta_k + \mu_k,$$
$$\mathbf{E}\left[\zeta_{k+1} \mid \mathcal{G}_k\right] \leq (1 - \theta_k)\zeta_k - \bar{u}_k + \epsilon_k \xi_k + \nu_k,$$

*where $\mathcal{G}_k$ denotes the collection $\{\xi_0, \ldots, \xi_k, \zeta_0, \ldots, \zeta_k, u_0, \ldots, u_k, \bar{u}_0, \ldots, \bar{u}_k, \eta_0, \ldots, \eta_k, \theta_0, \ldots, \theta_k, \epsilon_0, \ldots, \epsilon_k, \mu_0, \ldots, \mu_k, \nu_0, \ldots, \nu_k\}$, and $c$ is a positive scalar. Also, assume that*

$$\sum_{k=0}^{\infty} \eta_k < \infty, \qquad \sum_{k=0}^{\infty} \epsilon_k < \infty, \qquad \sum_{k=0}^{\infty} \mu_k < \infty, \qquad \sum_{k=0}^{\infty} \nu_k < \infty, \qquad w.p.1.$$

*Then $\xi_k$ and $\zeta_k$ converge almost surely to two nonnegative random variables, respectively, and we have*

$$\sum_{k=0}^{\infty} u_k < \infty, \qquad \sum_{k=0}^{\infty} \bar{u}_k < \infty, \qquad \sum_{k=0}^{\infty} \theta_k \zeta_k < \infty, \qquad w.p.1.$$

Now we analyze the $y$-step in the basic SCGD Algorithm 1. The motivation of introducing the auxiliary variable $y_{k+1}$ is to track the unknown quantity $g(x_k)$ by averaging past samples $\{g_{w_t}(x_t)\}_{t=0}^{k}$. We show in the next lemma that the distance between $y_{k+1}$ and $g(x_k)$ decreases "in expectation" according to a supermartingle-type inequality. Intuitively, this suggests that the tracking variable $y_{k+1}$ and the unknown $g(x_k)$ "converge" to each other. It is worth pointing out that this result does not require the convexity of the objective function.

**Lemma 2** *Let Assumption 1 hold, and let $\{(x_k, y_k)\}$ be generated by Algorithm 1. Then:*

(i) *With probability 1, we have*

$$\mathbf{E}\left[\|y_{k+1} - g(x_k)\|^2 \mid \mathcal{F}_k\right] \leq (1 - \beta_k)\|y_k - g(x_{k-1})\|^2 + \beta_k^{-1} C_g \|x_k - x_{k-1}\|^2 + 2V_g \beta_k^2. \quad (5)$$

(ii) *If $\sum_{k=1}^{\infty} \alpha_k^2 \beta_k^{-1} < \infty$, we have*

$$\sum_{k=0}^{\infty} \beta_k^{-1} C_g \|x_k - x_{k-1}\|^2 < \infty, \qquad w.p.1.$$

(iii) *There exists a scalar $D_y \geq 0$ such that $\mathbf{E}\left[\|y_{k+1} - g(x_k)\|^2\right] \leq D_y$ for all $k$.*

*Proof.* (i) Let $e_k = (1 - \beta_k)(g(x_k) - g(x_{k-1}))$. Together with the definition of $y_k$ [cf. Alg. 1], we have

$$y_{k+1} - g(x_k) + e_k = (1 - \beta_k)(y_k - g(x_{k-1})) + \beta_k(g_{w_k}(x_k) - g(x_k)). \quad (6)$$

In addition, using Assumption 1(iii), we obtain

$$\|e_k\| \leq (1 - \beta_k)\sqrt{C_g}\|x_k - x_{k-1}\|. \quad (7)$$

Taking squared norm expectation of both sides of Eq. (6) conditioned on $\mathcal{F}_k$ and using Assumption 1(ii)-(iii), we have

$$\begin{aligned}
\mathbf{E}\left[\|y_{k+1} - g(x_k) + e_k\|^2 \mid \mathcal{F}_k\right] &= \mathbf{E}\left[\|(1 - \beta_k)(y_k - g(x_{k-1})) + \beta_k(g_{w_k}(x_k) - g(x_k))\|^2 \mid \mathcal{F}_k\right] \\
&= (1 - \beta_k)^2 \|y_k - g(x_{k-1})\|^2 + \beta_k^2 \mathbf{E}\left[\|g_{w_k}(x_k) - g(x_k)\|^2 \mid \mathcal{F}_k\right] \\
&\quad + 2(1 - \beta_k)\beta_k (y_k - g(x_{k-1}))' \mathbf{E}\left[(g_{w_k}(x_k) - g(x_k)) \mid \mathcal{F}_k\right] \\
&\leq (1 - \beta_k)^2 \|y_k - g(x_{k-1})\|^2 + \beta_k^2 V_g + 0.
\end{aligned} \quad (8)$$

Utilizing the fact $\|a + b\|^2 \leq (1 + \epsilon)\|a\|^2 + (1 + 1/\epsilon)\|b\|^2$ for any $\epsilon > 0$, we have

$$\|y_{k+1} - g(x_k)\|^2 \leq (1 + \beta_k)\|y_{k+1} - g(x_k) + e_k\|^2 + (1 + 1/\beta_k)\|e_k\|^2.$$

Taking expectation of both sides and applying Eqs. (7)-(8), we obtain

$$\begin{aligned}
&\mathbf{E}\left[\|y_{k+1} - g(x_k)\|^2 \mid \mathcal{F}_k\right] \\
&\leq (1 + \beta_k)(1 - \beta_k)^2 \|y_k - g(x_{k-1})\|^2 + (1 + \beta_k)\beta_k^2 V_g + \frac{1 - \beta_k^2}{\beta_k} C_g \|x_k - x_{k-1}\|^2 \\
&\leq (1 - \beta_k)\|y_k - g(x_{k-1})\|^2 + \frac{C_g}{\beta_k}\|x_k - x_{k-1}\|^2 + 2V_g \beta_k^2,
\end{aligned}$$

for all $k$, with probability 1.

(ii) Under the additional assumption $\sum_{k=1}^{\infty} \alpha_k^2 \beta_k^{-1} < \infty$, we have

$$\sum_{k=0}^{\infty} \beta_k^{-1} \mathbf{E}\left[\|x_k - x_{k-1}\|^2\right] = \sum_{k=0}^{\infty} \mathcal{O}(\alpha_k^2 \beta_k^{-1}) < \infty,$$

where the equality holds by Eq. (4). By using the monotone convergence theorem, we obtain that $\sum_{k=0}^{T} \beta_k^{-1} C_g \|x_k - x_{k-1}\|^2$ converges almost surely to a random variable with finite expectation as $T \to \infty$. Therefore, $\sum_{k=0}^{\infty} \beta_k^{-1} C_g \|x_k - x_{k-1}\|^2$ exists and is finite w.p.1.

(iii) By taking $D_y = \mathbf{E}\left[\|y_1 - g(x_0)\| \right] + 2C_{\text{var}} + C_g^2 C_f \geq \mathbf{E}\left[\|y_1 - g(x_0)\| \right] + 2C_{\text{var}} \beta_k + C_g^2 C_f (\alpha_k / \beta_k)^2$ for all $k$, we can prove by induction that $\mathbf{E}\left[\|y_{k+1} - g(x_k)\|^2\right] \leq D_y$ for all $k$. ∎

Next we analyze the improvement of the optimality error $\|x_k - x^*\|$ as the algorithm proceeds. As we show in the following lemma, the optimality improvement is also coupled with the estimation error $\|g(x_k) - y_{k+1}\|$.

**Lemma 3** *Let Assumption 1 hold, and let $F = f \circ g$ be convex. Then the basic SCGD Algorithm 1 generates a sequence $\{(x_k, y_k)\}$ such that, with probability 1*

$$\mathbf{E}\left[\|x_{k+1} - x^*\|^2 \mid \mathcal{F}_k\right] \leq \left(1 + L_f^2 C_g \beta_k^{-1} \alpha_k^2\right) \|x_k - x^*\|^2 - 2\alpha_k \left(F(x_k) - F^*\right) + C_f C_g \alpha_k^2 \\ + \beta_k \mathbf{E}\left[\|g(x_k) - y_{k+1}\|^2 \mid \mathcal{F}_k\right]. \tag{9}$$

*Proof.* Let $x^*$ be an arbitrary optimal solution of problem (1), and let $F^* = F(x^*)$. Using the definition of $x_{k+1}$, the fact $\Pi_\mathcal{X} x^* = x^*$ and the nonexpansiveness of $\Pi_\mathcal{X}$, we have

$$\begin{aligned}
&\|x_{k+1} - x^*\|^2 \\
&= \left\|\Pi_\mathcal{X}\left\{x_k - x^* - \alpha_k \widetilde{\nabla} g_{w_k}(x_k) \nabla f_{v_k}(y_{k+1})\right\}\right\|^2 \\
&\leq \|x_k - x^* - \alpha_k \widetilde{\nabla} g_{w_k}(x_k) \nabla f_{v_k}(y_{k+1})\|^2 \\
&= \|x_k - x^*\|^2 + \alpha_k^2 \|\widetilde{\nabla} g_{w_k}(x_k) \nabla f_{v_k}(y_{k+1})\|^2 - 2\alpha_k (x_k - x^*)' \widetilde{\nabla} g_{w_k}(x_k) \nabla f_{v_k}(y_{k+1}) \\
&= \|x_k - x^*\|^2 + \alpha_k^2 \|\widetilde{\nabla} g_{w_k}(x_k) \nabla f_{v_k}(y_{k+1})\|^2 - 2\alpha_k (x_k - x^*)' \widetilde{\nabla} g_{w_k}(x_k) \nabla f_{v_k}(g(x_k)) + u_k,
\end{aligned}$$

where we define $u_k$ to be

$$u_k = 2\alpha_k (x_k - x^*)' \widetilde{\nabla} g_{w_k}(x_k) \left(\nabla f_{v_k}(g(x_k)) - \nabla f_{v_k}(y_{k+1})\right). \tag{10}$$

Taking expectation of both sides, using Assumption 1(ii) and Eq. (4), we obtain

$$\begin{aligned}
&\mathbf{E}\left[\|x_{k+1} - x^*\|^2 \mid \mathcal{F}_k\right] \\
&\leq \|x_k - x^*\|^2 + C_f C_g \alpha_k^2 - 2\alpha_k (x_k - x^*)' \mathbf{E}\left[\widetilde{\nabla} g_{w_k}(x_k) \nabla f_{v_k}(g(x_k)) \mid \mathcal{F}_k\right] + \mathbf{E}\left[u_k \mid \mathcal{F}_k\right].
\end{aligned}$$

By using Assumption 1(ii) and the convexity of $F = f \circ g$, we obtain

$$(x_k - x^*)' \mathbf{E}\left[\widetilde{\nabla} g_{w_k}(x_k) \nabla f_{v_k}(g(x_k)) \mid \mathcal{F}_k\right] \geq F(x_k) - F^*.$$

Then we have

$$\mathbf{E}\left[\|x_{k+1} - x^*\|^2 \mid \mathcal{F}_k\right] \leq \|x_k - x^*\|^2 + C_f C_g \alpha_k^2 - 2\alpha_k \left(F(x_k) - F^*\right) + \mathbf{E}\left[u_k \mid \mathcal{F}_k\right]. \tag{11}$$

Let us consider the $u_k$ term. By using norm inequalities and Assumption 1(iii)-(iv), we have

$$\begin{aligned}
u_k &\leq 2\alpha_k \|x_k - x^*\| \|\widetilde{\nabla} g_{w_k}(x_k)\| \|\nabla f_{v_k}(g(x_k)) - \nabla f_{v_k}(y_{k+1})\| \\
&\leq 2\alpha_k L_f \|x_k - x^*\| \|\widetilde{\nabla} g_{w_k}(x_k)\| \|g(x_k) - y_{k+1}\| \\
&\leq \beta_k \|g(x_k) - y_{k+1}\|^2 + \frac{\alpha_k^2}{\beta_k} L_f^2 \|x_k - x^*\|^2 \|\widetilde{\nabla} g_{w_k}(x_k)\|^2,
\end{aligned}$$

and by taking expectation as well as using Assumption 1(iv), we further have

$$\mathbf{E}\left[u_k \mid \mathcal{F}_k\right] \leq \beta_k \mathbf{E}\left[\|g(x_k) - y_{k+1}\|^2 \mid \mathcal{F}_k\right] + \frac{\alpha_k^2}{\beta_k} L_f^2 C_g \|x_k - x^*\|^2.$$

We apply the preceding inequality to Eq. (11), yielding

$$\mathbf{E}\left[\|x_{k+1} - x^*\|^2 \mid \mathcal{F}_k\right] \\ \leq \left(1 + L_f^2 C_g \frac{\alpha_k^2}{\beta_k}\right) \|x_k - x^*\|^2 - 2\alpha_k \left(F(x_k) - F^*\right) + C_f C_g \alpha_k^2 + \beta_k \mathbf{E}\left[\|g(x_k) - y_{k+1}\|^2 \mid \mathcal{F}_k\right]$$

and completing the proof. ∎

So far we have characterized the behaviors of the sequences $\{y_{k+1} - g(x_k)\}$ and $\{x_k - x^*\}$ in Lemma 2 and Lemma 3, respectively. Next we consider the behavior of $\{F(x_k)\}$. The following result will be useful when we apply the basic SCGD to nonconvex optimization. When $F$ is not necessarily convex, it is expected that the algorithm converges to some stationary point $\bar{x}$ of problem (1), i.e., a point $\bar{x}$ such that $\nabla F(x) = 0$. Let us assume for now that $F$ is differentiable and $\mathcal{X} = \Re^n$.

**Lemma 4** *Suppose that Assumption 1 holds, $F$ has Lipschitz continuous gradient with parameter $L_F > 0$, and $\mathcal{X} = \Re^n$. Then, if $\frac{\alpha_k}{\beta_k} L_f^2 C_g < 1/2$, Algorithm 1 generates a sequence $\{(x_k, y_k)\}$ such that*

$$\mathbf{E}\left[F(x_{k+1}) \mid \mathcal{F}_k\right] \leq F(x_k) - \frac{\alpha_k}{2}\|\nabla F(x_k)\|^2 + \alpha_k^2 L_F C_f C_g + \beta_k \mathbf{E}\left[\|g(x_k) - y_{k+1}\|^2 \mid \mathcal{F}_k\right], \quad (12)$$

*with probability 1.*

*Proof.* Using the fact that $F$ has Lipschitz continuous gradient, we have

$$\begin{aligned} F(x_{k+1}) &\leq F(x_k) + \nabla F(x_k)'(x_{k+1} - x_k) + L_F\|x_{k+1} - x_k\|^2 \\ &= F(x_k) - \alpha_k\|\nabla F(x_k)\|^2 + \alpha_k \nabla F(x_k)'(\nabla F(x_k) - \widetilde{\nabla} g_{w_k}(x_k)\nabla f_{v_k}(y_{k+1})) \\ &\quad + L_F\|x_{k+1} - x_k\|^2. \end{aligned} \quad (13)$$

We obtain that, by using norm inequalities, with probability 1

$$\begin{aligned} &\mathbf{E}\left[\nabla F(x_k)'(\nabla F(x_k) - \widetilde{\nabla} g_{w_k}(x_k)\nabla f_{v_k}(y_{k+1})) \mid \mathcal{F}_k\right] \\ &= \alpha_k \nabla F(x_k)' \mathbf{E}\left[\widetilde{\nabla} g_{w_k}(x_k)\left(\nabla f_{v_k}(g(x_k)) - \nabla f_{v_k}(y_{k+1})\right) \mid \mathcal{F}_k\right] \\ &\leq \alpha_k L_f \|\nabla F(x_k)\| \mathbf{E}\left[\|\widetilde{\nabla} g_{w_k}(x_k)\|\|g(x_k) - y_{k+1}\| \mid \mathcal{F}_k\right] \\ &\leq \beta_k \mathbf{E}\left[\|g(x_k) - y_{k+1}\|^2 \mid \mathcal{F}_k\right] + \frac{\alpha_k^2}{\beta_k} L_f^2 \|\nabla F(x_k)\|^2 \mathbf{E}\left[\|\widetilde{\nabla} g_{w_k}(x_k)\|^2 \mid \mathcal{F}_k\right] \\ &\leq \beta_k \mathbf{E}\left[\|g(x_k) - y_{k+1}\|^2 \mid \mathcal{F}_k\right] + \frac{\alpha_k^2}{\beta_k} L_f^2 C_g \|\nabla F(x_k)\|^2. \end{aligned}$$

Also note that $\mathbf{E}\left[\|x_{k+1} - x_k\|^2 \mid \mathcal{F}_k\right] \leq \alpha_k^2 C_f C_g$. Taking expectation of both sides of (13) and applying the preceding results, we have

$$\begin{aligned} &\mathbf{E}\left[F(x_{k+1}) \mid \mathcal{F}_k\right] \\ &\leq F(x_k) - \alpha_k \left(1 - \frac{\alpha_k}{\beta_k} L_f^2 C_g\right) \|\nabla F(x_k)\|^2 + \alpha_k^2 L_F C_f C_g + \beta_k \mathbf{E}\left[\|g(x_k) - y_{k+1}\|^2 \mid \mathcal{F}_k\right], \end{aligned}$$

which concludes the proof. ∎

## 2.2 Almost Sure Convergence

Now we prove the almost sure convergence of the basic SCGD (cf. Algorithm 1). A convergence proof for a special case of the algorithm has been given in [10]. In what follows, we present a generalized proof of the almost sure convergence, based on a coupled martingale convergence argument. Specifically, we show that the basic SCGD is guaranteed to find an optimal solution for convex optimization problems if such a solution exists. We also show that any limit point generated by the basic SCGD must be stationary for problems that are not necessarily convex. This proof of convergence sheds light on the convergence rate and error sample complexity analysis which we develop later.

**Theorem 5 (Almost Sure Convergence of basic SCGD)** *Let Assumption 1 hold, and let the stepsizes $\{\alpha_k\}$ and $\{\beta_k\}$ be such that*

$$\sum_{k=0}^{\infty} \alpha_k = \infty, \qquad \sum_{k=0}^{\infty} \beta_k = \infty, \qquad \sum_{k=0}^{\infty} \left(\alpha_k^2 + \beta_k^2 + \frac{\alpha_k^2}{\beta_k}\right) < \infty.$$

*Let $\{(x_k, y_k)\}$ be the sequence generated by the SCGD Algorithm 1 starting with an arbitrary initial point $(x_0, y_0)$. Then:*

(a) *If $F$ is convex and problem (1) has at least one optimal solution, $x_k$ converges almost surely to an optimal solution of problem (1).*

(b) *If $F$ has Lipschitz continuous gradient and $\mathcal{X} = \Re^n$, any limit point of the sequence $\{x_k\}$ is a stationary point with probability 1.*

*Proof.* (a) By Lemma 2, we have

$$\sum_{k=0}^{\infty} \frac{C_g}{\beta_k} \|x_k - x_{k-1}\|^2 < \infty, \qquad w.p.1.$$

This, together with the assumption $\sum_{k=0}^{\infty}(\alpha_k^2 + \beta_k^2 + \alpha_k^2/\beta_k) < \infty$ and the fact $F(x_k) - F^* \geq 0$, suggests that the supermartingale convergence Lemma 1 applies to Eqs. (5) and (9).

By applying Lemma 1, we obtain that $\{\|x_k - x^*\|\}$ and $\{\|y_{k+1} - g(x_k)\|\}$ converge almost surely to two random variables, and

$$\sum_{k=0}^{\infty} \alpha_k \left(F(x_k) - F^*\right) < \infty, \qquad \sum_{k=0}^{\infty} \beta_k \|y_{k+1} - g(x_k)\|^2 < \infty, \qquad w.p.1,$$

further implying that

$$\liminf_{k \to \infty} F(x_k) = F^*, \qquad \liminf_{k \to \infty} \|y_{k+1} - g(x_k)\|^2 = 0, \qquad w.p.1.$$

Since $\{\|x_k - x^*\|\}$ converges almost surely, the sequence $\{x_k\}$ is bounded with probability 1. Consider an arbitrary sample trajectory of $x_k$. By the continuity of $F$, the sequence $\{x_k\}$ must have a limit point $\bar{x}$ being an optimal solution, i.e., $F(\bar{x}) = F^*$. Since the choice of $x^*$ is arbitrary, we take $x^* = \bar{x}$ and $\|x_k - \bar{x}\| \to 0$. Then $x_k \to \bar{x}$ on this sample trajectory. Therefore $x_k$ converges almost surely to a random point in the set of optimal solutions of problem (1).

(b) We apply supermartingale convergence Lemma 1 to Eqs. (5) and (12). It follows that $\{F(x_k)\}$ and $\{\|y_{k+1} - g(x_k)\|\}$ converge almost surely to two random variables, and

$$\sum_{k=0}^{\infty} \alpha_k \|\nabla F(x_k)\|^2 < \infty, \qquad \sum_{k=0}^{\infty} \beta_k \|y_{k+1} - g(x_k)\|^2 < \infty, \qquad w.p.1.$$

We focus on a single sample trajectory of $x_k$ such that the preceding inequalities hold. Let $\epsilon > 0$ be arbitrary. We note that as long as $x_k$ has a limit point, then there exists at least one limit point, denoted as $\bar{x}$, satisfying $\|\nabla F(\bar{x})\| \leq \epsilon$. Otherwise we would have $\sum_{k=0}^{\infty} \alpha_k \|\nabla F(x_k)\|^2 \geq \sum_{k=0}^{\infty} \alpha_k \epsilon^2 = \infty$, yielding a contradiction. Consequently, there exists a set $\bar{N}$ (union of neighborhoods of all $\epsilon$-stationary limit points) such that $x_k$ visits infinitely often, and

$$\|\nabla F(x_k)\| \begin{cases} \leq \epsilon & \text{if } x_k \in \bar{N}, \\ \geq \epsilon & \text{if } x_k \notin \bar{N}, \end{cases}$$

for all $k$ sufficiently large.

We assume to the contrary that there exists a limit point $\widetilde{x}$ such that $\|\nabla F(\widetilde{x})\| > 2\epsilon$ for $x \in N(\widetilde{x})$. Then there exists a set $\widetilde{N}$ such that $x_k$ visits infinitely often, and

$$\|\nabla F(x_k)\| \begin{cases} \geq 2\epsilon & \text{if } x_k \in \widetilde{N}, \\ \leq 2\epsilon & \text{if } x_k \notin \widetilde{N}, \end{cases}$$

for all $k$ sufficiently large. Moreover, the sets $\bar{N}$ and $\widetilde{N}$ are disjoint. Since the sequence $x_k$ enters both $\bar{N}$ and $\widetilde{N}$ infinitely often, there exists a subsequence

$$\{x_k\}_{k \in \mathcal{K}} = \{x_{s_i}, x_{s_i+1}, \ldots, x_{t_i}\}_{i=1}^{\infty}$$

that crosses the two neighborhoods infinitely often. In other words, we have for every $i$ sufficiently large that

$$\|\nabla F(x_{t_i})\| \geq 2\epsilon > \|\nabla F(x_k)\| \geq \epsilon, \qquad \forall\, k = s_i, \ldots, t_i - 1,$$

and

$$x_{s_i} \in \bar{N}, \qquad x_{t_i} \in \widetilde{N}.$$

On one hand, by using the triangle inequality, we have

$$\sum_{k \in \mathcal{K}} \|x_{k+1} - x_k\| = \sum_{i=1}^{\infty} \sum_{k=s_i}^{t_i-1} \|x_{k+1} - x_k\| \geq \sum_{i=1}^{\infty} \|x_{t_i} - x_{s_i}\| \geq \sum_{i=1}^{\infty} \text{dist}(N(\bar{x}), N(\widetilde{x})) = \infty.$$

On the other hand, we have

$$\infty > \sum_{k=0}^{\infty} \alpha_k \|\nabla f(x_k)\|^2 \geq \sum_{k \in \mathcal{K}} \alpha_k \|\nabla f(x_k)\|^2 \geq \epsilon^2 \sum_{k \in \mathcal{K}} \alpha_k.$$

However, we can further obtain that $\sum_{k \in \mathcal{K}} \mathbf{E}\left[\|x_{k+1} - x_k\|\right] = \mathcal{O}\left(\sum_{k \in \mathcal{K}} \alpha_k\right) < \infty$. This together with the monotone convergence theorem implies that, on this subsequence $\mathcal{K}$,

$$\sum_{k \in \mathcal{K}} \|x_{k+1} - x_k\| < \infty, \qquad w.p.1.$$

This yields a contradiction with the sample path result $\sum_{k \in \mathcal{K}} \|x_{k+1} - x_k\| = \infty$ we just obtained. Therefore there does not exist a limit point $\widetilde{x}$ such that $\|\nabla F(\widetilde{x})\| > 2\epsilon$. Since $\epsilon$ can be made arbitrarily small, there does not exist any limit point that is nonstationary. In other words, any limit point of $x_k$ is a stationary point of $F(x)$. ∎

## 2.3 Rate of Convergence

Now we establish the convergence rate and sample complexity of the basic SCGD algorithm. To do this, we consider the averaged iterates of Algorithm 1, given by

$$\widehat{x}_k = \frac{1}{N_k} \sum_{t=k-N_k}^{k} x_t,$$

where we take $N_k = \lceil k/2 \rceil$ for simplicity. Note that the convergence rate is related to the stepsizes of the algorithm. By convention, we choose stepsizes $\{\alpha_k\}$ and $\{\beta_k\}$ as powers of $k$, i.e.,

$$\alpha_k = k^{-a}, \qquad \beta_k = k^{-b},$$

and we aim to minimize the error bound over the parameters $a, b$.

In what follows, we analyze three cases separately: (i) the case where $F = f \circ g$ is a convex function; (ii) the special case where $F = f \circ g$ is a strongly convex function; (iii) the case where $F = f \circ g$ is not necessarily convex. For the convex and strongly convex cases (i) and (ii), we consider the rate of convergence of Algorithm 1 in terms of the optimality error $F(x_k) - F^*$ and the distance to optimal solution $\|x_k - x^*\|$, respectively. For the nonconvex case (iii), we consider the convergence rate in terms of a metric of nonstationarity.

**Theorem 6 (Convergence Rate of basic SCGD for Convex Problems)** *Suppose that Assumption 1 holds, $F$ is convex, and there exists at least one optimal solution $x^*$ to problem (1). Let $D_x > 0$ be the diameter of the feasible set such that $\sup_{x \in \mathcal{X}} \|x - x^*\|^2 \leq D_x$, and let $D_y > 0$ be the scalar defined in Lemma 2. Let the stepsizes be*

$$\alpha_k = k^{-a}, \qquad \beta_k = k^{-b},$$

*where $a, b$ are scalars in $(0, 1)$. Then the averaged iterates generated by Algorithm 1 is such that*

$$\mathbf{E}\left[F(\widehat{x}_k) - F^*\right] = \mathcal{O}\Big((1 + L_f^2 C_g^2)(D_x + D_y)(k^{a-1} + k^{b-a}) + C_f C_g k^{-a} + V_g k^{a-2b} + C_g^2 C_f k^{b-a}\Big).$$

*For $k$ sufficiently large, minimizing the preceding bound with respect to $a$ and $b$ yields*

$$a^* = 3/4, \qquad b^* = 1/2,$$

*and the bound becomes*

$$\mathbf{E}\left[F(\widehat{x}_k) - F^*\right] = \mathcal{O}\Big(\frac{(1 + L_f^2 C_g^2)D + V_g + C_g^2 C_f}{k^{1/4}} + \frac{C_f C_g}{k^{3/4}}\Big).$$

*Proof.* Define the random variable

$$J_k = \|x_k - x^*\|^2 + \|y_k - g(x_{k-1})\|^2,$$

so we have $\mathbf{E}[J_k] \leq D_x + D_y \equiv D$ for all $k$. We multiply Eq. (5) by $(1 + \beta_k)$ and take its sum with Eq. (9), and we obtain

$$\mathbf{E}\left[J_{k+1} \mid \mathcal{F}_k\right] \leq \Big(1 + L_f^2 C_g^2 \frac{\alpha_k^2}{\beta_k}\Big) J_k - 2\alpha_k \left(F(x_k) - F^*\right)$$
$$+ \left(C_f C_g \alpha_k^2 + 2V_g \beta_k^2\right) + \frac{C_g}{\beta_k} \|x_k - x_{k-1}\|^2.$$

Taking expectation of both sides yields

$$\mathbf{E}\left[J_{k+1}\right] \leq \Big(1 + L_f^2 C_g^2 \frac{\alpha_k^2}{\beta_k}\Big) \mathbf{E}\left[J_k\right] - 2\alpha_k \mathbf{E}\left[F(x_k) - F^*\right] + C_f C_g \alpha_k^2 + 2V_g \beta_k^2 + C_g^2 C_f \frac{\alpha_k^2}{\beta_k}.$$

Let $N > 0$. By reordering the terms in the preceding relation and taking its sum over $k - N, \ldots, k$,

we have

$$2\sum_{t=k-N}^{k} \mathbf{E}\left[F(x_t) - F^*\right]$$

$$\leq \sum_{t=k-N}^{k} \frac{1}{\alpha_t}\left(\left(1 + L_f^2 C_g^2 \frac{\alpha_t^2}{\beta_t}\right)\mathbf{E}\left[J_t\right] - \mathbf{E}\left[J_{t+1}\right]\right) + \sum_{t=k-N}^{k}\left(C_f C_g \alpha_t + 2V_g \frac{\beta_t^2}{\alpha_t} + C_g^2 C_f \frac{\alpha_t}{\beta_t}\right)$$

$$= \sum_{t=k-N}^{k}\left(\frac{1}{\alpha_t} - \frac{1}{\alpha_{t-1}}\right)\mathbf{E}\left[J_t\right] - \frac{1}{\alpha_k}\mathbf{E}\left[J_{k+1}\right] + \frac{1}{\alpha_{k-N-1}}\mathbf{E}\left[J_{k-N}\right] + L_f^2 C_g^2 \sum_{t=k-N}^{k} \frac{\alpha_t}{\beta_t}\mathbf{E}\left[J_t\right]$$

$$+ C_f C_g \sum_{t=k-N}^{k} \alpha_t + 2V_g \sum_{t=k-N}^{k} \frac{\beta_t^2}{\alpha_t} + C_g^2 C_f \sum_{t=k-N}^{k} \frac{\alpha_t}{\beta_t}$$

$$\leq \sum_{t=k-N}^{k}\left(\frac{1}{\alpha_t} - \frac{1}{\alpha_{t-1}}\right)D + \frac{1}{\alpha_{k-N-1}}D + L_f^2 C_g^2 \sum_{t=k-N}^{k} \frac{\alpha_t}{\beta_t}D$$

$$+ C_f C_g \sum_{t=k-N}^{k} \alpha_t + 2V_g \sum_{t=k-N}^{k} \frac{\beta_t^2}{\alpha_t} + C_g^2 C_f \sum_{t=k-N}^{k} \frac{\alpha_t}{\beta_t}$$

$$\leq \frac{1}{\alpha_k}D + L_f^2 C_g^2 \left(\sum_{t=k-N}^{k} \frac{\alpha_t}{\beta_t}\right)D + C_f C_g \sum_{t=k-N}^{k} \alpha_t + 2V_g \sum_{t=k-N}^{k} \frac{\beta_t^2}{\alpha_t} + C_g^2 C_f \sum_{t=k-N}^{k} \frac{\alpha_t}{\beta_t}.$$

Let $\alpha_k = k^{-a}$, and $\beta_k = k^{-b}$, where $a, b$ are scalars in $(0, 1]$. We have

$$2\sum_{t=k-N}^{k} \mathbf{E}\left[F(x_t) - F^*\right]$$

$$\leq \mathcal{O}\bigg(k^a D + L_f^2 C_g^2 \left(k^{1+b-a} - (k-N)^{1+b-a}\right)D + C_f C_g(k^{1-a} - (k-N)^{1-a})$$

$$+ 2V_g(k^{1+a-2b} - (k-N)^{1+a-2b}) + C_g^2 C_f(k^{1+b-a} - (k-N)^{1+b-a})\bigg).$$

Using the convexity of $F$ and taking $N = N_k = k/2$, we obtain

$$\mathbf{E}\left[F(\widehat{x}_k) - F^*\right] \leq \frac{1}{N_k}\sum_{t=k-N_k}^{k} \mathbf{E}\left[F(x_t) - F^*\right]$$

$$\leq \mathcal{O}\left((k^{a-1} + L_f^2 C_g^2 k^{b-a})D + C_f C_g k^{-a} + V_g k^{a-2b} + C_g^2 C_f k^{b-a}\right).$$

The order of the bound is minimized when $a = 3/4$ and $b = 1/2$, which completes the proof. ∎

Next we consider a special case where $F = f \circ g$ is strongly convex in the following sense: there exists a scalar $\sigma > 0$ such that

$$F(x) - F^* \geq \sigma\|x - x^*\|^2, \qquad \forall\, x \in \mathcal{X}. \tag{14}$$

In the next theorem we show that a faster convergence rate can be obtained assuming strong convexity. This is consistent with the well known complexity results for convex optimization.

**Theorem 7 (Convergence Rate of basic SCGD for Strongly Convex Problems)** *Suppose that Assumption 1 holds and F is strongly convex satisfying* (14). *If the stepsizes are chosen as*

$$\alpha_k = \frac{1}{k\sigma}, \quad \beta_k = \frac{1}{(k\sigma)^{2/3}},$$

*we have*

$$\mathbf{E}\left[\|x_k - x^*\|^2\right] = \mathcal{O}\left(\frac{C_f C_g}{\sigma^2} \frac{\log k}{k} + \frac{L_f^2 C_g (C_f C_g^2 + V_g)}{\sigma^{8/3}} \frac{1}{k^{2/3}}\right) = \mathcal{O}\left(\frac{1}{k^{2/3}}\right),$$

*and*

$$\mathbf{E}\left[\|\widehat{x}_k - x^*\|^2\right] = \mathcal{O}\left(\frac{1}{k^{2/3}}\right).$$

*Proof.* We follow the line of analysis as in Lemma 3. We first derive a different bound for the $u_k$ term given by Eq. (10). This time we have

$$\mathbf{E}[u_k|\mathcal{F}_k] \leq \alpha_k \sigma \|x_k - x^*\|^2 + \frac{\alpha_k L_f^2 C_g}{\sigma} \mathbf{E}\left[\|g(x_k) - y_{k+1}\|^2 \mid \mathcal{F}_k\right]. \tag{15}$$

Plugging Eqs. (14) and (15) into Eq. (11), and taking expectation of both sides, we obtain

$$\mathbf{E}\left[\|x_{k+1} - x^*\|^2 \mathcal{F}_k\right] \leq (1 - \sigma \alpha_k) \mathbf{E}\left[\|x_k - x^*\|^2\right] + C_f C_g \alpha_k^2 + \frac{\alpha_k L_f^2 C_g}{\sigma} \mathbf{E}\left[\|g(x_k) - y_{k+1}\|^2\right]. \tag{16}$$

Next, we define the variable

$$J_k = \|x_k - x^*\|^2 + \Lambda(\alpha_k, \beta_k) \|g(x_{k-1}) - y_k\|^2,$$

where $\Lambda(\alpha_k, \beta_k) = \max\left\{\frac{L_f^2 C_g \alpha_k (\beta_k - 1)/\sigma}{\sigma \alpha_k - \beta_k}, 0\right\} = \Theta\left(\frac{L_f^2 C_g \alpha_k}{\beta_k \sigma}\right)$.

Multiplying Eq. (5) by $\Lambda(\alpha_k, \beta_k)$ and taking its sum with Eq. (16), we obtain

$$\mathbf{E}[J_{k+1}] \leq (1 - \sigma \alpha_k) \mathbf{E}[J_k] + C_f C_g \alpha_k^2 + \mathcal{O}\left(\frac{L_f^2 C_g}{\sigma}\right)\left(C_f C_g^2 \frac{\alpha_k^3}{\beta_k^2} + V_g \alpha_k \beta_k\right).$$

Taking $\alpha_k = 1/(\sigma k)$, $\beta_k = \alpha_k^{2/3}$ and multiplying the preceding inequality with $k$ on both sides, we have

$$k\mathbf{E}[J_{k+1}] \leq (k-1)\mathbf{E}[J_k] + \frac{C_f C_g}{k\sigma^2} + \mathcal{O}\left(\frac{L_f^2 C_g (C_f C_g^2 + V_g)}{k^{2/3} \sigma^{8/3}}\right).$$

Applying the preceding inequality inductively from 0 to $k$, we have

$$k\mathbf{E}[J_{k+1}] \leq \mathcal{O}\left(\frac{C_f C_g}{\sigma^2} \log k + \frac{L_f^2 C_g (C_f C_g^2 + V_g)}{\sigma^{8/3}} k^{1/3}\right).$$

Finally, we have

$$\mathbf{E}\left[\|x_{k+1} - x^*\|^2\right] \leq \frac{1}{k} \mathbf{E}[J_{k+1}] \leq \mathcal{O}\left(\frac{C_f C_g \log k}{\sigma^2 k} + \frac{L_f^2 C_g (C_f C_g^2 + V_g)}{\sigma^{8/3} k^{2/3}}\right) = \mathcal{O}(k^{-2/3}).$$

By the convexity of $\|\cdot\|$, the averaged iterates $\widehat{x}_k$ satisfy the same bound. ∎

Let us compare the rate of convergence for convex and strongly convex problems after $k$ queries to the $\mathcal{SO}$. For convex problems, the error is on the order of $k^{-1/4}$; while for strongly convex problems, the error is on the order of $k^{-2/3}$. It is as expected that strongly convex problems are "easier" to solve than those problems lacking strong convexity.

Finally we analyze the behavior of the basic SCGD for nonconvex optimization problems. Without convexity, the algorithm is no longer guaranteed to find a global optimum. However, we have shown that any limit point of iterates produced by the algorithm must be a stationary point for the nonconvex problem. In the next theorem, we provide an estimate that quantifies how fast the non-stationary metric $\|\nabla F(x_k)\|$ decrease to zero.

**Theorem 8 (Convergence Rate of basic SCGD for Nonconvex Problems)** *Suppose Assumption 1 holds, $F$ is differentiable, and $\mathcal{X} = \Re^n$. Let the stepsizes be*

$$\alpha_k = k^{-a}, \qquad \beta_k = k^{-b},$$

*where $a, b$ are scalars in $(0, 1)$. Let*

$$T_\epsilon = \min\left\{k : \inf_{0 \leq t \leq k} \mathbf{E}\left[\|\nabla F(x_t)\|^2\right] \leq \epsilon\right\}$$

*then*

$$T_\epsilon \leq \mathcal{O}(\epsilon^{-1/p}),$$

*where $p = \min\{1 - a, a - b, 2b - a, a\}$. By minimizing the complexity bound over $a, b$, we obtain $T_\epsilon \leq \mathcal{O}(\epsilon^{-4})$ with $a = 3/4, b = 1/2$.*

*Proof.* Let us define the random variable

$$J_k = F(x_k) + \|y_k - g(x_{k-1})\|^2.$$

Multiplying Eq. (5) with $(1 + \beta_k)$ and take its sum with Eq. (12), we have

$$\mathbf{E}[J_{k+1}|\mathcal{F}_k] \leq J_k - (\alpha_k/2)\|\nabla F(x_k)\|^2 + 2\beta_k^{-1}C_g\|x_k - x_{k+1}\|^2 + 4V_g\beta_k^2 + \alpha_k^2 L_f C_f C_g.$$

We follow the same line of analysis as given by Theorem 6. By using a similar argument, we can show that

$$\frac{1}{k}\sum_{t=1}^{k} \mathbf{E}[\|\nabla F(x_t)\|^2] \leq \mathcal{O}\big(k^{a-1}J_0 + k^{b-a}C_f^2 C_g + 4V_g k^{a-2b} + k^{-a}L_f C_f C_g\big) \leq \mathcal{O}(k^{-p}),$$

where $p(a, b) = \min\{1 - a, a - b, 2b - a, a\}$. By using the definition of $T_\epsilon$, we have

$$\mathbf{E}[\|\nabla F(x_k)\|^2] > \epsilon, \qquad \text{if } k < T_\epsilon.$$

Combining the preceding two relations, we obtain

$$\epsilon \leq \frac{1}{T_\epsilon}\sum_{k=1}^{T_\epsilon} \mathbf{E}[\|\nabla F(x_k)\|^2] \leq \mathcal{O}(T_\epsilon^{-p}),$$

implying that $T_\epsilon \leq \mathcal{O}(\epsilon^{-1/p})$. ∎

## 3 Acceleration for Smooth Convex Optimization

In this section, we propose an accelerated version of SCGD that achieves faster rate of convergence when the objective function is differentiable, which we refer to as *accelerated SCGD*. Recall our optimization problem

$$\min_{x \in \mathcal{X}} \ (f \circ g)(x),$$

where
$$g(x) = \mathbf{E}\left[g_w(x)\right], \qquad f(y) = \mathbf{E}\left[f_v(y)\right], \qquad \forall\, x \in \Re^n,\ y \in \Re^m.$$

In Section 2, we have obtained the convergence rate for basic SCGD in the case where $g_w$ may be nondifferentiable. In this section, we will show this performance can be improved if we restrict to smooth problems where both $f_v$ and $g_w$ are differentiable. We state the *accelerated stochastic compositional gradient descent* (accelerated SCGD) method as follows.

---

**Algorithm 2** Accelerated SCGD Algorithm

---

**Input:** $x_0 \in \Re^n, y_0 \in \Re^m$, stepsizes $\{\alpha_k\}, \{\beta_k\} \subset (0,1]$, $\mathcal{SO}$.
**Output:** The sequence $\{x_k\}$.

1: **for** $k = 0, 1, 2, \ldots$ **do**
2:     Query $\mathcal{SO}$ for gradients of $g, f$ at $x_k, y_k$ respectively, obtaining $\nabla g_{w_k}(x_k)$ and $\nabla f_{v_k}(y_k)$.
3:     Update
$$x_{k+1} = \Pi_{\mathcal{X}}\left[x_k - \alpha_k \nabla g_{w_k}(x_k) \nabla f_{v_k}(y_k)\right],$$
$$z_{k+1} = -\left(\frac{1}{\beta_k} - 1\right)x_k + \frac{1}{\beta_k}x_{k+1}.$$

4:     Query $\mathcal{SO}$ for the sample value of $g$ at $z_{k+1}$, obtaining $g_{w_{k+1}}(z_{k+1})$.
5:     Update
$$y_{k+1} = (1 - \beta_k)y_k + \beta_k g_{w_{k+1}}(z_{k+1}).$$

6: **end for**

---

In Algorithm 2, $x_k, z_k \in \Re^n$, $y_k \in \Re^m$, $\nabla g_w$ is the $n \times m$ matrix with each column being the gradient of the corresponding entry of $g_w$, and $\{\alpha_k\}, \{\beta_k\}$ are pre-specified sequences of positive scalars in $(0, 1]$.

Note that the iteration for $x_k$ in Algorithm 2 takes the same form as in Algorithm 1. The auxiliary variable $z_{k+1}$ in Algorithm 2 is a query point obtained by extrapolation. Equivalently, the variable $x_{k+1}$ can be viewed as an interpolated point given by

$$x_{k+1} = (1 - \beta_k)x_k + \beta_k z_{k+1}.$$

Compared with the basic SCGD, the new estimate $y_k$ is a weighted average of $\{g_{w_t}(z_t)\}_{t=0}^k$, which are samples of function $g(\cdot)$ at the extrapolated points. Our purpose is to find an "unbiased" approximation of $g(x_k)$ in the following sense:

$$y_k = \text{weighted average of } \{g_{w_t}(z_t)\}_{t=0}^k \quad \approx \quad g(x_k) = g\big(\text{weighted average of } \{z_t\}_{t=0}^k\big).$$

In Algorithm 2, the new estimate $y_k$ "tracks" the unknown quantity $g(x_k)$ at a faster rate. As a result, the accelerated algorithm achieves an error complexity of $\mathcal{O}\left(k^{-2/7}\right)$ for general convex problems and $\mathcal{O}\left(k^{-4/5}\right)$ for strongly convex problems.

The acceleration of our algorithm comes from the extrapolation step that calculates $z_k$. Note that this idea is fundamentally different from the accelerated SA [12, 13]. The accelerated SA uses the idea of Nesterov smoothing [24] from deterministic convex optimization to improve the error dependency of SA on model parameters. In contrast, our accelerated SCGD solves the two-level stochastic problem (1), and it uses an extrapolation technique to balance noises associated with random samples, which reduces bias in the estimates and improves the error dependency on the number of samples $k$.

### 3.1 Almost Sure Convergence

In addition to Assumption 1, we make the following assumption.

**Assumption 2** *The function g has Lipschitz continuous gradients, i.e., there exists a scalar $L_g > 0$ such that*
$$\|\nabla g(x) - \nabla g(y)\| \leq L_g \|x - y\|, \qquad \forall\, x, y \in \mathcal{X}.$$
*The sample gradients of g and f have bounded fourth moments with probability 1, i.e.,*
$$\mathbf{E}\left[\|\nabla g_{w_0}(x)\|^4 \mid v_0\right] \leq C_g^2, \quad \mathbf{E}[\|\nabla f_{v_0}(y)\|^4 \mid w_0] \leq C_f^2, \qquad \forall\, x \in \mathcal{X},\ y \in \Re^m.$$

Now we state the almost sure convergence result of the accelerated SCGD. We will establish it through a series of lemmas.

**Theorem 9 (Almost Sure Convergence of the Accelerated SCGD)** *Suppose Assumptions 1 and 2 hold. Let the stepsizes $\{\alpha_k\}$ and $\{\beta_k\}$ satisfy*
$$\sum_{k=0}^{\infty} \alpha_k = \infty, \qquad \sum_{k=0}^{\infty} \beta_k = \infty, \qquad \sum_{k=0}^{\infty} \left(\alpha_k^2 + \beta_k^2 + \frac{\alpha_k^2}{\beta_k} + \frac{\alpha_k^4}{\beta_k^3}\right) < \infty.$$

*Let $\{(x_k, y_k)\}$ be generated by the accelerated SCGD Algorithm 2 starting from an arbitrary initial solution $(x_0, y_0)$. Then*

*(a) If F is convex and problem (1) has at least one optimal solution, $\{x_k\}$ converges almost surely to an optimal solution.*

*(b) If F has Lipschitz continuous gradient and $\mathcal{X} = \Re^n$, any limit point of $\{x_k\}$ is a stationary point with probability 1.*

The key step of the convergence proof is to characterize the behavior of $y_k - g(x_k)$. To do so, we attempt to write $y_k$ as a weighted average of samples at the extrapolated points $\{g_{w_t}(z_t)\}_{t=0}^k$. Let us define the weights $\xi_t^{(k)}$ as
$$\xi_t^{(k)} = \begin{cases} \beta_t \prod_{i=t+1}^{k}(1 - \beta_i) & \text{if } k > t \geq 0, \\ \beta_k & \text{if } k = t \geq 0. \end{cases}$$

Without loss of generality, we take $\beta_0 = 1$ throughout. By the definitions of $\xi_t^{(k)}$, $x_k$ and $y_k$, we observe
$$\xi_t^{(k+1)} = (1 - \beta_{k+1})\xi_t^{(k)}, \qquad \sum_{t=0}^{k} \xi_t^{(k)} = 1, \qquad x_{k+1} = \sum_{t=0}^{k} \xi_t^{(k)} z_{t+1}, \qquad y_{k+1} = \sum_{t=0}^{k} \xi_t^{(k)} g_{w_{t+1}}(z_{t+1}).$$

Let us summarize a few basic inequalities which hold under Assumptions 1 and 2. We denote by $\mathcal{F}_k$ the collection of random variables
$$\{x_0, \ldots, x_k, y_0, \ldots, y_{k-1}, z_0, \ldots, z_k, w_0, \ldots, w_{k-1}, v_0, \ldots, v_{k-1}\}.$$

We first have, by Assumption 2 and the nonexpansiveness of projection,
$$\mathbf{E}\left[\|z_{k+1} - x_k\|^4 \mid \mathcal{F}_k\right] \leq \alpha_k^4 \beta_k^{-4} \mathbf{E}\left[\|\nabla g_{w_k}(x_k)\|^4 \|\nabla f_{v_k}(y_k)\|^4 \mid \mathcal{F}_k\right] \leq \alpha_k^4 \beta_k^{-4} C_g^2 C_f^2.$$

Since $x_{k+1} = (1 - \beta_k)x_k + \beta_k z_{k+1}$, we have
$$x_{k+1} - z_{k+1} = (1 - \beta_k)(x_k - z_{k+1}), \qquad x_{k+1} - x_k = \beta_k(z_{k+1} - x_k).$$

It follows that
$$\mathbf{E}\left[\|x_{k+1} - z_{k+1}\|^4 \mid \mathcal{F}_k\right] = \mathbf{E}\left[(1 - \beta_k)^4 \|x_k - z_{k+1}\|^4 \mid \mathcal{F}_k\right] \leq \alpha_k^4 \beta_k^{-4} C_g^2 C_f^2,$$

$$\mathbf{E}\left[\|x_{k+1} - x_k\|^4 \mid \mathcal{F}_k\right] = \mathbf{E}\left[\beta_k^4 \|z_{k+1} - x_k\|^4 \mid \mathcal{F}_k\right] \leq \alpha_k^4 C_g^2 C_f^2,$$

and by the Jensen's inequality,

$$\mathbf{E}\left[\|x_{k+1} - x_k\|^2 \mid \mathcal{F}_k\right] \leq \sqrt{\mathbf{E}\left[\|x_{k+1} - x_k\|^4 \mid \mathcal{F}_k\right]} \leq \alpha_k^2 C_g C_f.$$

Moreover, by Assumption 2 that the function $g$ has Lipschitz continuous gradient, we have

$$\|g(y) - g(x) + \nabla g(y)'(x - y)\| \leq \frac{L_g}{2}\|x - y\|^2, \qquad \forall\, x, y \in \mathcal{X}. \tag{17}$$

In what follows, we prove Theorem 9 through a series of lemmas. These lemmas are also the building blocks for the subsequent accelerated rate of convergence analysis. Our first lemma decomposes the estimation error $y_{k+1} - g(x_{k+1})$ into a sum of squared "biases" and a linear function of the past zero-mean "noises".

**Lemma 10 (Decomposition of $y_k - g(x_k)$)** *Let Assumptions 1 and 2 hold. Then*

$$\|y_{k+1} - g(x_{k+1})\| \leq L_g \sum_{t=0}^{k} \xi_t^{(k)} \|x_{k+1} - z_{t+1}\|^2 + \left\|\sum_{t=0}^{k} \xi_t^{(k)} (g_{w_{t+1}}(z_{t+1}) - g(z_{t+1}))\right\|.$$

*Proof.* We have

$$\begin{aligned} y_{k+1} &= \sum_{t=0}^{k} \xi_t^{(k)} g_{w_{t+1}}(z_{t+1}) = \sum_{t=0}^{k} \xi_t^{(k)} g(z_{t+1}) + \sum_{t=0}^{k} \xi_t^{(k)} (g_{w_{t+1}}(z_{t+1}) - g(z_{t+1})) \\ &= \sum_{t=0}^{k} \xi_t^{(k)} \left(g(x_{k+1}) + \nabla g(x_{k+1})'(z_{t+1} - x_{k+1}) + \zeta(x_{k+1}, z_{t+1})\right) + \sum_{t=0}^{k} \xi_t^{(k)} (g_{w_{t+1}}(z_{t+1}) - g(z_{t+1})) \\ &= \left(\sum_{t=0}^{k} \xi_t^{(k)}\right) g(x_{k+1}) + \nabla g(x_{k+1})' \sum_{t=0}^{k} \xi_t^{(k)} (z_{t+1} - x_{k+1}) + \sum_{t=0}^{k} \xi_t^{(k)} \zeta(x_{k+1}, z_{t+1}) \\ &\quad + \sum_{t=0}^{k} \xi_t^{(k)} (g_{w_{t+1}}(z_{t+1}) - g(z_{t+1})), \end{aligned} \tag{18}$$

where the second equality holds by the Taylor expansion of $g(\cdot)$ at $x_{k+1}$ that

$$g(z_{t+1}) = g(x_{k+1}) + \nabla g(x_{k+1})'(z_{t+1} - x_{k+1}) + \zeta(x_{k+1}, z_{t+1}),$$

with $\zeta(x_{k+1}, z_{t+1})$ summarizing the second and higher order terms. A *critical observation* here is that the first order term in Eq. (18) cancels out, i.e.,

$$\nabla g(x_{k+1})' \sum_{t=0}^{k} \xi_t^{(k)}(z_{t+1} - x_{k+1}) = \nabla g(x_{k+1})' \left(\sum_{t=0}^{k} \xi_t^{(k)} z_{t+1} - \left(\sum_{t=0}^{k} \xi_t^{(k)}\right) x_{k+1}\right) = 0,$$

where we use the facts that $x_{k+1} = \sum_{t=0}^{k} \xi_t^{(k)} z_{t+1}$ and $\sum_{t=0}^{k} \xi_t^{(k)} = 1$. Next, it follows from Eq. (18) that

$$y_{k+1} = g(x_{k+1}) + 0 + \sum_{t=0}^{k} \xi_t^{(k)} \zeta(x_{k+1}, z_{t+1}) + \sum_{t=0}^{k} \xi_t^{(k)} (g_{w_{t+1}}(z_{t+1}) - g(z_{t+1})).$$

Using triangle inequality, we have

$$\|y_{k+1} - g(x_{k+1})\| \leq \sum_{t=0}^{k} \xi_t^{(k)} \|\zeta(x_{k+1}, z_{t+1})\| + \left\|\sum_{t=0}^{k} \xi_t^{(k)}(g_{w_{t+1}}(z_{t+1}) - g(z_{t+1}))\right\|$$

$$\leq L_g \sum_{t=0}^{k} \xi_t^{(k)} \|x_{k+1} - z_{t+1}\|^2 + \left\|\sum_{t=0}^{k} \xi_t^{(k)}(g_{w_{t+1}}(z_{t+1}) - g(z_{t+1}))\right\|,$$

where the second inequality uses $\|\zeta(x_{k+1}, z_{t+1})\| \leq L_g \|x_{k+1} - z_{t+1}\|^2$ from Eq. (17). ■

In Lemma 10, we have bounded the error $\|y_k - g(x_k)\|$ by the sum of two iterative sums. In the next lemma, we consider the convergence properties of the two iterative sums.

**Lemma 11** Let Assumptions 1 and 2 hold, and let $q_{k+1} = \sum_{t=0}^{k} \xi_t^{(k)} \|x_{k+1} - z_{t+1}\|$, $m_{k+1} = \sum_{t=0}^{k} \xi_t^{(k)} \|x_{k+1} - z_{t+1}\|^2$, and $n_{k+1} = \sum_{t=0}^{k} \xi_t^{(k)}(g_{w_{t+1}}(z_{t+1}) - g(z_{t+1}))$. Then with probability 1,

(a) $q_{k+1}^2 \leq (1 - \beta_k)q_k^2 + (4/\beta_k)\|x_{k+1} - x_k\|^2$.

(b) $m_{k+1} \leq (1 - \beta_k)m_k + \beta_k q_k^2 + (2/\beta_k)\|x_k - x_{k+1}\|^2$.

(c) $\mathbf{E}\left[\|n_{k+1}\|^2 \mid \mathcal{F}_{k+1}\right] \leq (1 - \beta_k)^2 \|n_k\|^2 + \beta_k^2 V_g$.

*Proof.* (a) By the definition of $q_k$ and $\xi_t^{(k)}$, we have

$$q_{k+1} = \sum_{t=0}^{k} \xi_t^{(k)} \|x_{k+1} - z_{t+1}\|$$

$$= (1 - \beta_k) \sum_{t=0}^{k-1} \xi_t^{(k-1)} \|x_{k+1} - z_{t+1}\| + \beta_k \|x_{k+1} - z_{k+1}\|$$

$$\leq (1 - \beta_k) \sum_{t=0}^{k-1} \xi_t^{(k-1)} (\|x_k - z_{t+1}\| + \|x_{k+1} - x_k\|) + \beta_k \|x_{k+1} - z_{k+1}\|$$

$$= (1 - \beta_k)q_k + (1 - \beta_k)\|x_{k+1} - x_k\| + \beta_k \|x_{k+1} - z_{k+1}\|$$

$$= (1 - \beta_k)q_k + 2(1 - \beta_k)\|x_{k+1} - x_k\|,$$

where the inequality holds by the triangle inequality, the third equality holds by the definition of $q_k$ and the fact $\sum_{t=0}^{k} \xi_t^{(k)} = 1$, and the last equality holds by the definition of $z_{k+1}$. Taking squares of both sides and using the inequality $(a+b)^2 \leq (1+\beta)a^2 + (1+1/\beta)b^2$ for any $\beta > 0$, we obtain

$$q_{k+1}^2 \leq (1+\beta_k)(1-\beta_k)^2 q_k^2 + (1+\beta_k^{-1})4(1-\beta_k)^2 \|x_{k+1} - x_k\|^2 \leq (1-\beta_k)q_k^2 + 4\beta_k^{-1}\|x_{k+1} - x_k\|^2.$$

(b) By the definition of $m_k$, we have

$$m_{k+1} = \sum_{t=0}^{k} \xi_t^{(k)} \|x_{k+1} - z_{t+1}\|^2 = (1-\beta_k)\sum_{t=0}^{k-1} \xi_t^{(k-1)} \|x_{k+1} - z_{t+1}\|^2 + \beta_k \|x_{k+1} - z_{k+1}\|^2$$

$$= (1-\beta_k)m_k + (1-\beta_k)\sum_{t=0}^{k-1} \xi_t^{(k-1)}\left(\|x_{k+1} - z_{t+1}\|^2 - \|x_k - z_{t+1}\|^2\right) + \beta_k \|x_{k+1} - z_{k+1}\|^2.$$

Using the triangle inequality, we have

$$\|x_{k+1} - z_{t+1}\|^2 - \|x_k - z_{t+1}\|^2 = (\|x_{k+1} - z_{t+1}\| - \|x_k - z_{t+1}\|)(\|x_{k+1} - z_{t+1}\| + \|x_k - z_{t+1}\|)$$

$$\leq \|x_{k+1} - x_k\|(2\|x_k - z_{t+1}\| + \|x_{k+1} - x_k\|).$$

It follows that

$$m_{k+1} \leq (1-\beta_k)m_k + (1-\beta_k)\Big(2\|x_{k+1} - x_k\|\sum_{t=0}^{k-1}\xi_t^{(k-1)}\|x_k - z_{t+1}\| + \sum_{t=0}^{k-1}\xi_t^{(k-1)}\|x_{k+1} - x_k\|^2\Big)$$
$$+ \beta_k\|x_{k+1} - z_{k+1}\|^2$$
$$= (1-\beta_k)m_k + 2(1-\beta_k)\|x_{k+1} - x_k\|q_k + (1-\beta_k)\|x_{k+1} - x_k\|^2 + (1-\beta_k)^2/\beta_k\|x_{k+1} - x_k\|^2$$
$$\leq (1-\beta_k)m_k + (1-\beta_k)\Big(\beta_k^{-1}\|x_{k+1} - x_k\|^2 + \beta_k q_k^2\Big) + (1/\beta_k - 1)\|x_{k+1} - x_k\|^2$$
$$\leq (1-\beta_k)m_k + \beta_k q_k^2 + 2\beta_k^{-1}\|x_{k+1} - x_k\|^2,$$

where the first equality uses the definitions of $q_k, z_{k+1}$ and the fact $\sum_{t=0}^{k-1}\xi_t^{(k-1)} = 1$.

(c) By the definition of $n_k$, we have

$$n_{k+1} = (1-\beta_k)n_k + \beta_k\big(g_{w_{k+1}}(z_{k+1}) - g(z_{k+1})\big).$$

Taking conditional expectation of both sides and using the fact $n_k \in \mathcal{F}_{k+1}$, we further obtain

$$\mathbf{E}\big[\|n_{k+1}\|^2 \mid \mathcal{F}_{k+1}\big]$$
$$= (1-\beta_k)^2\|n_k\|^2 + 2\beta_k(1-\beta_k)n_k'\mathbf{E}\big[g_{w_{k+1}}(z_{k+1}) - g(z_{k+1}) \mid \mathcal{F}_{k+1}\big]$$
$$+ \beta_k^2\mathbf{E}\big[\|g_{w_{k+1}}(z_{k+1}) - g(z_{k+1})\|^2 \mid \mathcal{F}_{k+1}\big]$$
$$\leq (1-\beta_k)^2\|n_k\|^2 + \beta_k^2 V_g,$$

where we used the unbiasedness and moment boundedness of $g_w$ [Assumption 1(i),(ii)]. ∎

Now we use Lemma 10 and Lemma 11 to derive an iterative inequality for an upper bound of $\|y_k - g(x_k)\|$. The following lemma is an analogy of Lemma 2.

**Lemma 12** *Let Assumptions 1 and 2 hold. Then there exists a random variable $e_k \in \mathcal{F}_{k+1}$ for all $k$ satisfying $\|y_k - g(x_k)\| \leq e_k$ and*

$$\mathbf{E}\big[e_{k+1}^2 \mid \mathcal{F}_{k+1}\big] \leq \Big(1 - \frac{\beta_k}{2}\Big)e_k^2 + 2\beta_k^2 V_g + \mathcal{O}\Big(\frac{L_g^2}{\beta_k^3}\Big)\|x_k - x_{k+1}\|^4,$$

*with probability 1, and*

$$\mathbf{E}\big[e_1^2\big] \leq 2V_g, \qquad \mathbf{E}\big[e_{k+1}^2\big] \leq \Big(1 - \frac{\beta_k}{2}\Big)\mathbf{E}\big[e_k^2\big] + \mathcal{O}\Big(L_g^2 C_f^2 C_g^2 \frac{\alpha_k^4}{\beta_k^3} + V_g\beta_k^2\Big).$$

*Proof.* By Lemmas 10 and 11, we have

$$\|y_k - g(x_k)\|^2 \leq (L_g m_k + \|n_k\|)^2 \leq 2L_g^2 m_k^2 + 2\|n_k\|^2.$$

Let us we find an upper bound for $m_k^2$. In addition, by the iterative inequalities for $q_k$ and $m_k$ derived in Lemma 11 (a) and (b), we obtain that, for some $C > 0$ (e.g., $C = 4$),

$$m_{k+1} + Cq_{k+1}^2 \leq (1-\beta_k/2)(m_k + Cq_k^2) + \mathcal{O}\Big(\beta_k^{-1}\|x_k - x_{k+1}\|^2\Big).$$

Taking squares of the both sides of the above inequality and using the fact $(a+b)^2 \leq (1+\beta/2)a^2 + (1+2/\beta)b^2$, we have

$$\big(m_{k+1} + Cq_{k+1}^2\big)^2 \leq (1-\beta_k/2)\big(m_k + Cq_k^2\big)^2 + \mathcal{O}(\beta_k^{-3}\|x_k - x_{k+1}\|^4). \tag{19}$$

We define
$$e_k^2 = 2L_g^2 \left(m_k + Cq_k^2\right)^2 + 2\|n_k\|^2,$$
then clearly $\|y_k - g(x_k)\|^2 \leq e_k^2$ for all $k$. Taking the sum of Eq. (19) and the iterative inequality for $n_k$ derived in Lemma 11(c), we have
$$\mathbf{E}\left[e_{k+1}^2 \mid \mathcal{F}_{k+1}\right] \leq (1 - \beta_k/2)e_k^2 + 2\beta_k^2 V_g + \mathcal{O}(L_g^2 \beta_k^{-3})\|x_k - x_{k+1}\|^4.$$

Taking expectation on both sides, we have
$$\mathbf{E}\left[e_{k+1}^2\right] \leq (1 - \beta_k/2)\mathbf{E}\left[e_k^2\right] + \mathcal{O}(\beta_k^2 V_g + \beta_k^{-3}\alpha_k^4 L_g^2 C_g^2 C_f^2).$$

Since $\beta_0 = 1$, we have $x_1 = z_1$ and $m_1 = q_1 = 0$. In addition, by the fact that $\mathbf{E}\left[\|n_1\|^2\right] = V_g$, it holds that $\mathbf{E}\left[e_1^2\right] \leq 2V_g$, which concludes the proof. ∎

By now, we have shown the convergence property of $\|y_k - g(x_k)\|$ and its upper bound $e_k$. We are ready to prove Theorem 9.

**Proof of Theorem 9.** (a) Let $x^*$ be an arbitrary optimal solution of problem (1), and let $F^* = F(x^*)$. Since the iteration for calculating $x_k$ in Algorithm 1 has a similar form as the $x_k$ iteration for Algorithm 1, by the same argument in Lemma 3, we obtain
$$\mathbf{E}\left[\|x_{k+1} - x^*\|^2 \mid \mathcal{F}_k\right] \leq \|x_k - x^*\|^2 + \alpha_k^2 C_f C_g - 2\alpha_k(F(x_k) - F^*) + \mathbf{E}\left[u_k \mid \mathcal{F}_k\right],$$
where we define $u_k$ to be the vector
$$u_k = 2\alpha_k(x_k - x^*)'\nabla g_{w_k}(x_k)(\nabla f_{v_k}(g(x_k)) - \nabla f_{v_k}(y_k)).$$

We analyze $u_k$ in the same way as in Lemma 3, yielding that $\mathbf{E}[u_k \mid \mathcal{F}_k] \leq \beta_k \mathbf{E}\left[e_k^2 \mid \mathcal{F}_k\right] + \frac{\alpha_k^2}{\beta_k}L_f^2 C_g\|x_k - x^*\|^2$. It follows that
$$\mathbf{E}\left[\|x_{k+1} - x^*\|^2 \mid \mathcal{F}_k\right] \leq \left(1 + L_f^2 C_g \frac{\alpha_k^2}{\beta_k}\right)\|x_k - x^*\|^2 + \alpha_k^2 C_f C_g - 2\alpha_k(F(x_k) - F^*) + \beta_k \mathbf{E}\left[e_k^2 \mid \mathcal{F}_k\right]. \tag{20}$$

Recall that by Lemma 12, we have
$$\mathbf{E}\left[\mathbf{E}\left[e_{k+1}^2 \mid \mathcal{F}_{k+1}\right] \mid \mathcal{F}_k\right] \leq \left(1 - \frac{\beta_k}{2}\right)\mathbf{E}\left[e_k^2 \mid \mathcal{F}_k\right] + 2\beta_k^2 V_g + \mathcal{O}\left(\frac{\alpha_k^4}{\beta_k^3}L_g^2 C_f^2 C_g^2\right). \tag{21}$$

By the assumption that $\sum_{k=0}^{\infty}\left(\alpha_k^2 + \beta_k^2 + \frac{\alpha_k^2}{\beta_k} + \frac{\alpha_k^4}{\beta_k^3}\right) < \infty$, we can apply the coupled supermartingale Lemma 1 to inequalities (20) and (21). The remaining proof follows the same line as in Theorem 5(a). It is obtained that $x_k$ converges almost surely to an optimal solution.

(b) The $x_k$ update steps of Algorithm 1 and Algorithm 2 are equivalent. Thus, Lemma 4 holds for Algorithm 2 as well. It follows that
$$\mathbf{E}\left[F(x_{k+1}) \mid \mathcal{F}_k\right] \leq F(x_k) - \frac{\alpha_k}{2}\|\nabla F(x_k)\|^2 + \alpha_k^2 L_F C_f C_g + \beta_k \mathbf{E}\left[e_k^2 \mid \mathcal{F}_k\right],$$
with probability 1. We apply the supermartingale convergence argument to the preceding inequality and the inequality (21). The rest of the proof follows the same line of analysis as in Theorem 5(b). ∎

## 3.2 Accelerated Rate of Convergence

Next we analyze the rate of convergence for the accelerated SCGD Algorithm 2. We consider three cases: convex problems, strongly convex problems, nonconvex problems. Similar to Section 2, we also consider the averaged iterates given by

$$\widehat{x}_k = \frac{1}{N_k} \sum_{t=k-N_k}^{k} x_t,$$

where $N_k = \lceil k/2 \rceil$. The key component of our analysis is the recursive improvement in $\|y_k - g(x_k)\|$, which has been established in Lemma 12. As Lemma 12 gives a sharper bound compared with Lemma 2, the accelerated SCGD generates more accurate estimates of $g(x_k)$, resulting in faster rate of convergence.

**Theorem 13 (Accelerated Convergence Rate for General Convex Problems)** *Suppose that Assumptions 1 and 2 hold, $F$ is a convex function, and there exists at least one optimal solution $x^*$ to problem (1). Let $D > 0$ be a constant such that $\sup_{x \in \mathcal{X}} \|x - x^*\|^2 \leq D$, and let the stepsizes be*

$$\alpha_k = k^{-a}, \qquad \beta_k = k^{-b},$$

*where $a, b$ are scalars in $(0, 1)$. Then the averaged iterates generated by the accelerated SCGD Algorithm 2 is such that*

$$\mathbf{E}\left[F(\widehat{x}_k) - F^*\right] \leq \mathcal{O}\left(Dk^{a-1} + C_f C_g k^{-a} + C_1 k^{-b/2} + C_2 k^{-2a+2b}\right),$$

*where $C_1 = \sqrt{DC_g V_g} L_f$, and $C_2 = \sqrt{DC_g} L_f L_g C_g C_f$.*

*Proof.* By the proof of Theorem 9, we have

$$\mathbf{E}\left[\|x_{k+1} - x^*\|^2 \mid \mathcal{F}_k\right] \leq \|x_k - x^*\|^2 + \alpha_k^2 C_f C_g - 2\alpha_k(F(x_k) - F^*) + \mathbf{E}\left[u_k \mid \mathcal{F}_k\right],$$

where $u_k = 2\alpha_k(x_k - x^*)' \nabla g_{w_k}(x_k)(\nabla f_{v_k}(g(x_k)) - \nabla f_{v_k}(y_k))$. Taking expectation of both sides and reordering the terms, we obtain

$$2\mathbf{E}\left[F(x_k) - F^*\right] \leq \frac{1}{\alpha_k} \mathbf{E}\left[\|x_k - x^*\|^2 - \|x_{k+1} - x^*\|^2\right] + \alpha_k C_f C_g + \frac{1}{\alpha_k} \mathbf{E}\left[u_k\right].$$

Taking the sum of the preceding inequalities over $t = k - N, \ldots, k$, we have

$$2 \sum_{t=k-N}^{k} \mathbf{E}\left[F(x_t) - F^*\right] \leq \sum_{t=k-N}^{k} \frac{1}{\alpha_t}\left(\mathbf{E}\left[\|x_t - x^*\|^2\right] - \mathbf{E}\left[\|x_{t+1} - x^*\|^2\right]\right) + \sum_{t=k-N}^{k}\left(C_f C_g \alpha_t + \frac{1}{\alpha_t}\mathbf{E}\left[u_t\right]\right)$$

$$= \sum_{t=k-N}^{k} \left(\frac{1}{\alpha_t} - \frac{1}{\alpha_{t-1}}\right)\mathbf{E}\left[\|x_t - x^*\|^2\right] - \frac{1}{\alpha_k}\mathbf{E}\left[\|x_k - x^*\|^2\right]$$

$$+ \frac{1}{\alpha_{k-N-1}}\mathbf{E}\left[\|x_{k-N} - x^*\|^2\right] + C_f C_g \sum_{t=k-N}^{k} \alpha_t + \sum_{t=k-N}^{k} \frac{1}{\alpha_t}\mathbf{E}\left[u_t\right].$$

Using the fact $\|x_k - x^*\|^2 \leq D$, the Lipschitz continuity of $\nabla f$, and the Cauchy-Schwartz inequality, we obtain

$$\frac{1}{\alpha_k}\mathbf{E}\left[u_k\right] \leq 2\|x_k - x^*\|\mathbf{E}\left[\|\nabla g_{w_k}(x_k)\|\|\nabla f_{v_k}(y_k) - \nabla f_{v_k}(g(x_k))\|\right]$$

$$\leq 2\sqrt{D} L_f \mathbf{E}\left[\|\nabla g_{w_k}(x_k)\|^2\right]^{1/2} \mathbf{E}\left[\|y_k - g(x_k)\|^2\right]^{1/2}$$

$$\leq 2\sqrt{DC_g} L_f \mathbf{E}\left[e_k^2\right]^{1/2}.$$

Using $\|x_k - x^*\|^2 \leq D$ again, we further obtain

$$2 \sum_{t=k-N}^{k} \mathbf{E}\left[F(x_t) - F^*\right]$$

$$\leq \sum_{t=k-N}^{k} \left(\frac{1}{\alpha_t} - \frac{1}{\alpha_{t-1}}\right) D + \frac{1}{\alpha_{k-N-1}} D + C_f C_g \sum_{t=k-N}^{k} \alpha_t + 2\sqrt{DC_g} L_f \sum_{t=k-N}^{k} \mathbf{E}\left[e_t^2\right]^{1/2} \quad (22)$$

$$= \frac{D}{\alpha_k} + C_f C_g \sum_{t=k-N}^{k} \alpha_t + 2\sqrt{DC_g} L_f \sum_{t=k-N}^{k} \mathbf{E}\left[e_t^2\right]^{1/2}.$$

By taking $\alpha_k = k^{-a}$, $\beta_k = k^{-b}$, we find an upper bound for the quantity $\sum_{t=k-N}^{k} \mathbf{E}\left[e_t^2\right]^{1/2}$. Using Lemma 12, we have

$$\mathbf{E}\left[e_{k+1}^2\right] \leq (1 - k^{-b}/2)\mathbf{E}\left[e_k^2\right] + \mathcal{O}\left(V_g k^{-2b} + L_g^2 C_g^2 C_f^2 k^{-4a+3b}\right), \quad (23)$$

and we can easily prove by induction that for all $k$,

$$\mathbf{E}\left[e_{k+1}^2\right] \leq \mathcal{O}\left(V_g k^{-2b+1} + L_g^2 C_g^2 C_f^2 k^{-4a+3b+1}\right),$$

Letting $N = k/2$, we reorder Eq. (23) and take its sum over $t = k - N, \ldots, k$ to obtain

$$\frac{1}{N} \sum_{t=k-N}^{k} (t^{-b}/2) \mathbf{E}\left[e_t^2\right] \leq \frac{1}{N} \sum_{t=k-N}^{k} \left(\mathbf{E}\left[e_t^2\right] - \mathbf{E}\left[e_{t+1}^2\right] + \mathcal{O}\left(V_g t^{-2b} + L_g^2 C_g^2 C_f^2 t^{-4a+3b}\right)\right)$$

$$\leq \frac{1}{N} \mathbf{E}\left[e_{k-N}^2\right] + \frac{1}{N} \sum_{t=k-N}^{k} \mathcal{O}\left(V_g t^{-2b} + L_g^2 C_g^2 C_f^2 t^{-4a+3b}\right)$$

$$= \mathcal{O}\left(V_g k^{-2b} + L_g^2 C_g^2 C_f^2 k^{-4a+3b}\right).$$

Using the Cauchy-Schwartz inequality, we have

$$\frac{1}{N} \sum_{t=k-N}^{k} \mathbf{E}\left[e_t^2\right]^{1/2} \leq \left(\frac{1}{N} \sum_{t=k-N}^{k} t^b\right)^{1/2} \left(\frac{1}{N} \sum_{t=k-N}^{k} t^{-b} \mathbf{E}\left[e_t^2\right]\right)^{1/2} = \mathcal{O}\left(V_g^{1/2} k^{-b/2} + L_g C_g C_f k^{-2a+2b}\right).$$

Finally, we return to Eq. (22). Using the convexity of $F$ and applying the inequality above, we obtain

$$\mathbf{E}\left[F\left(\frac{1}{N_k} \sum_{t=k-N_k}^{k} x_t\right) - F^*\right] \leq \frac{1}{N_k} \sum_{t=k-N_k}^{k} \mathbf{E}\left[F(x_t) - F^*\right]$$

$$\leq Dk^{a-1} + C_f C_g \mathcal{O}\left(k^{-a}\right) + \sqrt{DC_g} L_f \mathcal{O}\left(V_g^{1/2} k^{-b/2} + L_g C_g C_f k^{-2a+2b}\right),$$

where the second inequality uses the fact $N_k = \Theta(k)$. ∎

To minimize the upper bound given by Theorem 13 for $k$ sufficiently large, we take

$$a^* = \frac{5}{7}, \qquad b^* = \frac{4a^*}{5} = \frac{4}{7}.$$

Then we have

$$\mathbf{E}\left[F(\widehat{x}_t) - F^*\right] \leq \mathcal{O}\left(\frac{D + \sqrt{DC_g V_g} L_f + \sqrt{DC_g} L_f L_g C_g C_f}{k^{2/7}} + \frac{C_f C_g}{k^{5/7}}\right) = \mathcal{O}(k^{-2/7}).$$

This proves acceleration as compared to the $\mathcal{O}(k^{-1/4})$ error bound of the basic SCGD as shown in Theorem 5.

Next we consider the case when the problem is strongly convex, i.e., there exists $\sigma > 0$ and a unique optimal solution $x^*$ such that $F(x) - F^* \geq \sigma\|x - x^*\|^2$ for all $x \in \mathcal{X}$.

**Theorem 14 (Accelerated Convergence Rate for Strongly Convex Problems)** *Let Assumptions 1 and 2 hold, and let $F$ be strongly convex with parameter $\sigma$. Let the stepsizes be*

$$\alpha_k = \frac{1}{\sigma k}, \qquad \beta_k = \frac{1}{(\sigma k)^{4/5}}.$$

*Then the iterates generated by Algorithm 2 is such that*

$$\mathbf{E}\left[\|x_k - x^*\|^2\right] \leq \mathcal{O}\Big(\frac{L_g C_f^2 C_g^2 + V_g}{\sigma^{14/5} k^{4/5}} + \frac{C_f C_g \log k}{\sigma^2 k}\Big) = \mathcal{O}(k^{-4/5}).$$

*Proof.* Recalling the proof of Theorem 9, we have

$$\mathbf{E}\left[\|x_{k+1} - x^*\|^2 \mid \mathcal{F}_k\right] \leq \|x_k - x^*\|^2 + \alpha_k^2 C_f C_g - 2\alpha_k(F(x_k) - F^*) + \mathbf{E}\left[u_k \mid \mathcal{F}_k\right], \qquad (24)$$

where

$$u_k = 2\alpha_k (x_k - x^*)' \nabla g_{w_k}(x_k)(\nabla f_{v_k}(g(x_k)) - \nabla f_{v_k}(y_k)).$$

By strong convexity, we have

$$F(x_k) - F^* \geq \sigma\|x_k - x^*\|^2.$$

By Assumption 1, we have

$$\mathbf{E}\left[u_k \mid \mathcal{F}_k\right] = 2\alpha_k (x_k - x^*)' \mathbf{E}\left[\nabla g_{w_k}(x_k)(\nabla f_{v_k}(g(x_k)) - \nabla f_{v_k}(y_k)) \mid \mathcal{F}_k\right]$$

$$\leq \alpha_k \sigma\|x_k - x^*\|^2 + \frac{L_f^2 C_g}{\sigma} \alpha_k \mathbf{E}\left[\|g(x_k) - y_k\|^2 \mid \mathcal{F}_k\right]$$

$$\leq \alpha_k \sigma\|x_k - x^*\|^2 + \frac{L_f^2 C_g}{\sigma} \alpha_k \mathbf{E}\left[e_k^2 \mid \mathcal{F}_k\right],$$

where the third relation uses the fact $e_k^2 \geq \|y_k - g(x_k)\|^2$ from Lemma 12.

Taking expectation of both sides of (24) and applying the preceding relations, we obtain

$$\mathbf{E}\left[\|x_{k+1} - x^*\|^2\right] \leq (1 - \sigma\alpha_k)\mathbf{E}\left[\|x_k - x^*\|^2\right] + C_f C_g \alpha_k^2 + \frac{L_f^2 C_g}{\sigma} \alpha_k \mathbf{E}\left[e_k^2\right]. \qquad (25)$$

Recalling Lemma 12, we have

$$\mathbf{E}\left[e_1^2\right] \leq 2V_g, \qquad \mathbf{E}\left[e_{k+1}^2\right] \leq \Big(1 - \frac{\beta_k}{2}\Big)\mathbf{E}\left[e_k^2\right] + \mathcal{O}\Big(L_g^2 C_f^2 C_g^2 \frac{\alpha_k^4}{\beta_k^3} + V_g \beta_k^2\Big). \qquad (26)$$

In what follows we analyze the convergence rate based on these two iterative inequalities. Let us define the variable

$$J_k = \|x_k - x^*\|^2 + \Lambda_k e_k^2.$$

where $\Lambda_k$ is a scalar given by

$$\Lambda_k = \max\left\{\frac{\alpha_k L_f^2 C_g}{\sigma(2^{-1}\beta_k - \sigma\alpha_k)}, 0\right\}.$$

By our choice of $\alpha_k$ and $\beta_k$, the scalar $\Lambda_k$ satisfies $0 \leq \Lambda_{k+1} \leq \Lambda_k = \frac{L_f^2 C_g}{\sigma}\Theta(\alpha_k \beta_k^{-1})$ for $k$ sufficiently large.

By applying Eq. (25), Eq. (26), and the properties of $\Lambda_k$, we obtain for $k$ sufficiently large that

$$\mathbf{E}\left[J_{k+1}\right] = \mathbf{E}\left[\|x_{k+1} - x^*\|^2 + \Lambda_{k+1} e_{k+1}^2\right] \leq \mathbf{E}\left[\|x_{k+1} - x^*\|^2 + \Lambda_k e_{k+1}^2\right]$$
$$\leq (1 - \sigma\alpha_k)\mathbf{E}\left[J_k\right] + C_f C_g \alpha_k^2 + \frac{L_f^2 C_g}{\sigma}\mathcal{O}\left(2\alpha_k \beta_k V_g + \alpha_k^5 \beta_k^{-4} L_g^2 C_g^2 C_f^2\right).$$

Taking $\alpha_k = (\sigma k)^{-1}$ and $\beta_k = \alpha_k^{4/5}$, we have for $k$ sufficiently large that

$$\mathbf{E}\left[J_{k+1}\right] \leq \left(1 - \frac{1}{k}\right)\mathbf{E}\left[J_k\right] + \frac{C_f C_g}{k^2 \sigma^2} + \mathcal{O}\left(\frac{L_f^2 C_g}{\sigma^{14/5}} \cdot \frac{V_g + L_g^2 C_g^2 C_f^2}{k^{9/5}}\right).$$

Multiplying both sides with $k$ and using induction, we obtain

$$k\mathbf{E}\left[J_{k+1}\right] \leq (k-1)\mathbf{E}\left[J_k\right] + \frac{C_f C_g}{k\sigma^2} + \mathcal{O}\left(\frac{L_f^2 C_g}{\sigma^{14/5}} \cdot \frac{V_g + L_g^2 C_g^2 C_f^2}{k^{4/5}}\right)$$
$$\leq \mathcal{O}\left(\frac{C_f C_g (\log k + 1)}{\sigma^2} + \frac{k^{1/5} L_f^2 C_g (V_g + L_g^2 C_g^2 C_f^2)}{\sigma^{14/5}}\right).$$

Therefore $\mathbf{E}\left[\|x_{k+1} - x^*\|^2\right] \leq \frac{1}{k}\mathbf{E}\left[J_{k+1}\right] \leq \mathcal{O}\left(\frac{C_f C_g \log k}{\sigma^2 k} + \frac{L_f^2 C_g (V_g + L_g^2 C_g^2 C_f^2)}{\sigma^{14/5} k^{4/5}}\right) = \mathcal{O}(k^{-4/5})$. ∎

Lastly, we consider the case where the objective function is not necessarily convex. We obtain the following convergence rate in terms of the nonstationarity metric $\mathbf{E}\left[\|\nabla F(x_k)\|^2\right]$.

**Theorem 15 (Convergence Rate of accelerated SCGD for Nonconvex Problems)** *Suppose that Assumptions 1 and 2 hold, $F$ has Lipschitz continuous gradient, and $\mathcal{X} = \Re^n$. Let the stepsizes be*

$$\alpha_k = k^{-5/7}, \qquad \beta_k = k^{-4/7}.$$

*Let $T_\epsilon = \min\left\{k : \inf_{0 \leq t \leq k} \mathbf{E}\left[\|\nabla F(x_t)\|^2\right] \leq \epsilon\right\}$, then $T_\epsilon \leq \mathcal{O}\left(\epsilon^{-7/2}\right).$*

The proof is analogous to Theorem 8, which we omit to avoid repetition.

## Remarks on Sample Error Complexity

An open issue remaining is whether the sample error complexity obtained by SCGD algorithms is improvable for problem (1)-(2). The best convergence rates we have obtained are $\mathcal{O}\left(k^{-2/7}\right)$ and $\mathcal{O}\left(k^{-4/5}\right)$ for convex and strongly convex problems, respectively. They are still inferior to the best known rates for the classical problem (3), i.e., $\mathcal{O}\left(k^{-1/2}\right)$ and $\mathcal{O}\left(1/k\right)$, which are known to be optimal in terms of $k$ (see [1]). However, the stochastic problem (1)-(2) considered here is more generic than problem (3), and in fact, contains the latter as a special case. We conjecture that, the stochastic problem (1)-(2) where one minimizes compositions of two expected-value functions is intrinsically harder than the expectation minimization problem (3). For future work, it will be very interesting to analyze the sample complexity lower bounds for the stochastic composition problem (1), either to show that the current rates are non-improvable, or to motivate faster algorithms.

# 4 Applications and Extensions

Optimization involving compositions of expected-value functions is a generic model of real-world decision making problems. In this section, we collect several instances of problem (1)-(2), and demonstrate the potential applications of the SCGD methods. We believe that the stochastic optimization model considered in this paper will motivate new applications as well as new approaches to tackle important practical problems.

## 4.1 Statistical Learning

We first demonstrate the application of SCGD to an important statistical learning problem on estimating sparse additive models (SpAM) [27]. SpAM is a class of models for high-dimensional sparse nonparametric regression. More specifically, suppose we are given a number of input vectors $\mathbf{x}_i = (x_{i1}, x_{i2}, ..., x_{id})^T \in \Re^d$ and responses $y_i$, where $i = 1, \ldots, n$. SpAM assumes that each paired sample $(\mathbf{x}_i, y_i)$ satisfies

$$y_i = \alpha + \sum_{j=1}^d h_j(x_{ij}) + \epsilon_i,$$

where each $h_j: \Re \to \Re$ is a feature component function, and each $\epsilon_i$ is some zero-mean noise. Without loss of generality, we always assume $\alpha = 0$ (This can be achieved by centering the data).

To make the model interpretable, SpAM assumes that most of the feature functions $h_j(\cdot)$ are zero functions. To induce sparsity, penalized regression methods have been well studied in the past decade (See [32, 35, 2] for examples), in which the component function $h_j$'s are assumed to be linear most of the time. When $h_j$'s are linear, the model simplifies to $y_i = \mathbf{x}_i^T \boldsymbol{\beta} + \epsilon_i$ for some $\boldsymbol{\beta} \in \Re^d$. In this case, $\ell_1$-penalized regression [32] is popularly known as the LASSO estimator, written as

$$\widehat{\boldsymbol{\beta}} = \mathrm{argmin}_{\boldsymbol{\beta} \in \mathbb{R}^d} \sum_{i=1}^n (y_i - \mathbf{x}_i^T \boldsymbol{\beta})^2 + \lambda \|\boldsymbol{\beta}\|_1.$$

The LASSO estimator has strong statistical guarantee and can be computed efficiently.

In nonparametric models, the feature functions $h_j$ are usually nonlinear and the $\ell_1$ penalty no longer induces sparsity. SpAM estimates the feature functions $h_j(\cdot)$ by solving the following stochastic minimization problem:

$$\min_{h_j \in \mathcal{H}_j, j=1\ldots,d} \mathbf{E}\Big[Y - \sum_{j=1}^d h_j(X_j)\Big]^2 + \lambda \sum_{j=1}^d \sqrt{\mathbf{E}\big[h_j^2(X_j)\big]}, \tag{27}$$

where $\mathcal{H}_j$ is some pre-specified functional space which ensures that the model is identifiable, and the expectation is taken over the sample pair $(X, Y)$. The SpAM estimator is known to have strong statistical guarantee [14, 27]. However, there has been no known stochastic algorithm to solve the SpAM problem efficiently. In particular, even though the problem in (27) is convex, a provably convergent stochastic algorithm for solving (27) is lacking. Since the SpAM estimator (27) takes the form of (1), we can apply the SCGD methods to solve the estimation problem (27). The stochastic nature of SCGD allows it to sequentially update the solution when a new pair of data point $(X, Y)$ arrives. We will provide a numerical example for solving SpAM using SCGD in Section 5.1.

## 4.2 Minimax Problems

The SCGD methods are suitable for online solution of the stochastic minimax problems

$$\min_{x \in \mathcal{X}} \max \big\{ \mathbf{E}\big[g_w^{(1)}(x)\big], \ldots, \mathbf{E}\big[g_w^{(I)}(x)\big] \big\},$$

where $I$ is a positive integer; $\mathbf{E}\big[g_w^{(i)}(x)\big]$ are convex functions. More generally, we may formulate the minimax problem to involve two stages of stochasticity, i.e.,

$$\min_{x \in \mathcal{X}} \mathbf{E}_v \left[\max\left\{\mathbf{E}\big[g_w^{(1)}(x) \mid v\big], \ldots, \mathbf{E}\big[g_w^{(I)}(x) \mid v\big]\right\}\right].$$

These problems can be viewed as stochastic games with random payoffs against an adversarial opponent. They can be seen to take the form of (1)-(2), if we consider the max operator as the outer function $f$. Note that the max operator is nondifferentiable. Thus we may use a smooth approximation to replace the max operator, then apply the SCGD methods to solve the approximate minimax problem.

### 4.3 Dynamic Programming

Dynamic programming is a rich area that involves sequential decision making, Monte Carlo estimation, and stochastic optimization. We consider the Markov decision problem (MDP) with states $i = 1, \ldots, n$. Finding an optimal policy for the MDP can be equivalently casted into solving the fixed-point *Bellman equation*, i.e., finding $J \in \Re^n$ such that

$$J = \max_{a \in A}\{g_a + \gamma P_a J\},$$

where $J \in \Re^n$ is the cost-to-go vector, $a \in A$ is an action, $g_a \in \Re^n$ is the transition cost vector given action $a$, $P_a \in \Re^{n \times n}$ is the matrix of transition probabilities given action $a$, $\gamma \in (0, 1]$ is a discount factor, and the maximization is taken elementwise. The optimal policy is the state to action mapping that achieves the maximization.

Meanwhile, the Bellman residual minimization approach is to solve the problem

$$\min_{J \in \Re^n} \|J - \max_a\{g_a + \gamma P_a J\}\|^2,$$

whose set of optimal solutions coincides with that of the Bellman equation. In *approximate dynamic programming*, we may solve the high-dimensional Bellman equation approximately by restricting $J$ to belong to some parametric family, which translates to a constraint in the residual minimization problem.

We consider the simulation setting where $P_a$ and $g_a$ are not explicitly given. Instead, we are given a simulator of the MDP in which, at the current state $x_k$, we can choose an arbitrary action $a_k$ to obtain a random transition cost $g_k$ and to arrive at a future state $i_{k+1}$. This generates a sample trajectory of state, action, and cost triplets according to the unknown transition probabilities:

$$\{(i_0, a_0, g_0), \ldots, (i_k, a_k, g_k), (i_{k+1}, a_{k+1}, g_{k+1}), \ldots\}.$$

We rewrite the Bellman residual minimization problem as

$$\min_{J \in \Re^n} \sum_{i=1}^n \Big(J(i) - \max_{a \in A}\big\{\mathbf{E}\left[g_k + \gamma J(i_{k+1}) \mid i_k = i, a_k = a\right]\big\}\Big)^2,$$

which takes the form of problem (1)-(2). Therefore we are able to apply SCGD to the Bellman minimization problem. The SCGD can update online based on the simulation trajectories $\{(i_k, a_k, g_k)\}$, without knowing the underlying transition probabilities. This suggests a new way to solve dynamic programming online.

## 4.4 Estimation of Rate Function

For complicated stochastic systems such as queues, estimation of certain tail probabilities is of both theoretical and practical interests. One example is to estimate the rate function of a random variable $X$ taking values in $\Re$, defined by

$$I(x) = -\lim_{n\to\infty} \frac{1}{n} \log \mathbf{P}\Big(\frac{1}{n}\sum_{i=1}^{n} X_i > x\Big), \qquad \forall\, x \in \Re,$$

where $X_i$ are i.i.d. samples of $X$. Finding the rate function $I(x)$ for a given $X$ can be casted into a convex optimization problem

$$I(x) = \sup_{\theta \in \Re} \big\{ \theta x - \log \mathbf{E}\left[\exp(\theta X)\right] \big\}.$$

which involves a nonlinear function of an expected-value function. Therefore the SCGD algorithm can be used for estimating $I(x)$ based on random realizations of $X$. It would be interesting to combine the optimization algorithm with adaptive Monte Carlo sampling to sharpen the sample complexity of the estimator.

## 4.5 Derivative-Free Optimization

An important extension of the SCGD is its application in derivative free optimization. Let us consider the setting where $f(x)$ is explicitly known, while $g(x)$ can be evaluated with noise but its gradient is not directly available. This is more general than the current setting where both the values and gradients of $g$ can be queried with random noise.

In the spirit of Kiefer-Wolfowitz method [15], we may approximate the gradient of $g$ with

$$\nabla g(x) \approx h(x) = \frac{g_w(x+d) - g_{\widetilde{w}}(x-d)}{\|d\|^2} d,$$

where $d$ is a randomly chosen vector, $w$ and $\widetilde{w}$ are independent variables.

By replacing the sample gradient $\widetilde{\nabla} g_w(x)$ in the Algorithm 1 with the approximate gradient, we obtain the derivative-free version of basic SCGD:

$$h_{k+1} = \frac{g_{w_k}(x_k + d_k) - g_{\widetilde{w}_k}(x_k - d_k)}{\|d_k\|^2} d_k$$
$$y_{k+1} = (1 - \beta_k) y_k + \beta_k g_{w_k}(x_k),$$
$$x_{k+1} = \Pi_{\mathcal{X}}\big\{ x_k - \alpha_k h_k \nabla f(y_{k+1}) \big\},$$

This differs from Algorithm 1 in two ways: (i) the gradient approximation of $g_k$ could be biased; and (ii) the sequence $\{d_k\}$ needs to be chosen in accordance with the stepsizes $\{\alpha_k\}$ and $\{\beta_k\}$. Similarly, we can obtain the derivative-free version of the accelerated SCGD, which we omit for simplicity. The derivative-free versions of SCGD potentially apply to a broader class of stochastic optimization problems. Analyzing the complexity of these algorithms and identifying their applications are interesting subjects for future research.

# 5 Numerical Results

In this section, we conduct numerical experiment of the SCGD methods for two problems.

## 5.1 Sparse Additive Model

We apply the SCGD algorithms to solve the SpAM optimization problem in (27). More specifically, let $\phi_1, \phi_2, \ldots,$ be a set of countable basis functions of the functional space $\mathcal{H}_j$'s, i.e., any function $h \in \mathcal{H}_j$ can be represented by $h(x) = \sum_{i=1}^{\infty} a_i \phi_i(x)$, where $a_i \in \Re$. In applications, we generally choose a number $p$ and approximate the function class $\mathcal{H}_j$ by

$$\widetilde{\mathcal{H}}_j^p := \Big\{ h : \text{there exist } a_1, \ldots, a_p \in \Re, \text{ such that } h(x) = \sum_{i=1}^{p} a_i \phi_i(x) \Big\}.$$

Using the above approximation, it is readily seen that problem (27) can be written into the form of problem (1). In particular, let

$$g(\eta) = \mathbf{E}\left[\left(\Big\{Y - \sum_{j=1}^{d}\sum_{k=1}^{p} \eta_{jk}\phi_k(X_k)\Big\}^2, \Big\{\sum_{k=1}^{p} \eta_{1k}\phi_1(X_1)\Big\}^2, \ldots, \Big\{\sum_{k=1}^{p} \eta_{dk}\phi_k(X_d)\Big\}^2\right)\right]^T,$$

where $\eta \in \Re^{d \times p}$, and

$$f(\mathbf{z}) = z_1 + \sum_{j=1}^{d} \sqrt{z_{j+1}}.$$

We see that the estimation problem (27) becomes $\min_\eta f(g(\eta))$. Estimating the feature functions $h_1, \ldots, h_d$ reduces to finding the weights $\eta$. The SCGD method is a natural fit for this stochastic optimization problem.

We simulate data from the model

$$y_i = \sum_{j=1}^{d} h_j(x_{ij}) + \epsilon_i,$$

where we choose the underlying feature functions as $h_1(t) = t^3$, $h_2(t) = t$, $h_3(t) = \cdots = h_d(t) = 0$ and $\epsilon_i \sim N(0, 0.1)$. Let $d = 30$ and $n = 200$. We use the span of cubic B-spline with 5 evenly distributed knots as the functional space $\widetilde{\mathcal{H}}_j = \text{span}\{\phi_1, \ldots, \phi_p\}$ (this choice of $\widetilde{\mathcal{H}}_j$ is consistent with [14]), and we choose the tuning parameter $\lambda$ by 3-fold cross-validation. Let all $\mathbf{x}_i$'s be independently drawn from the uniform distribution $U[0,1]^d$. We estimate the feature functions $h_j$'s by adopting basic SCGD Algorithm 1, where at the $k$-th iteration, we choose the stepsizes $\alpha_k = k^{-3/4}$ and $\beta_k = k^{-1/2}$. We run the algorithm for 20,000 iterations, and we plot the resulting solutions for $h_1$, $h_2$, $h_3$ and $h_4$ in Figure 1.

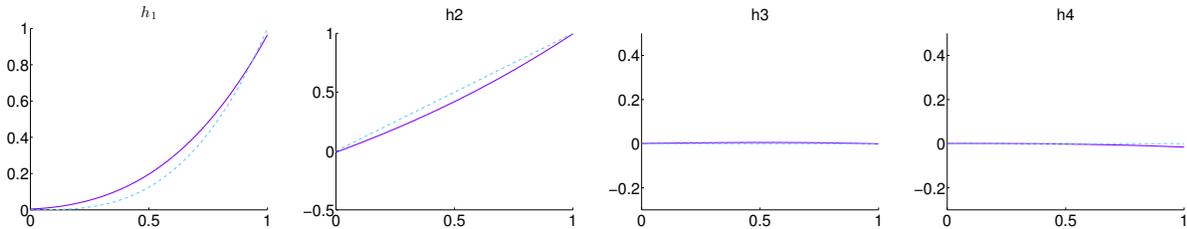

Figure 1: The top four feature functions $h_1, h_2, h_3, h_4$ of the SpAM model estimated by the basic SCGD. The feature functions $h_j$ are computed by solving (27) using basic SCGD based on random samples of input vector and response. The top four computed (purple solid lines) feature functions are plotted in the figure, compared against the true feature functions (blue doted lines).

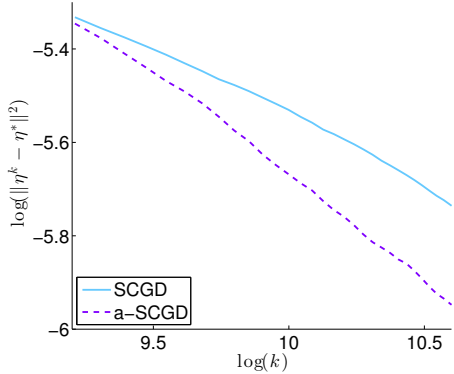

Figure 2: The empirical rate of convergences of basic and accelerated SCGD, where $\eta^k$ is the solution at the $k$-th iteration, and $\eta^*$ is the optimal solution. The accelerated SCGD converges at a faster rate compared with the basic SCGD.

We also investigate the empirical rate of convergence of the basic SCGD and the accelerated SCGD. Since the underlying true $\eta^*$ is unknown, we first let Algorithm 1 run $K = 1,000,000$ iterations and take the resulting solution as optimal $\eta^*$. Then, we run basic SCGD in Algorithm 1 with stepsizes $\alpha_k = k^{-3/4}$ and $\beta_k = k^{-1/2}$ and accelerated SCGD in Algorithm 2 with stepsizes $\alpha_k = k^{-5/7}$ and $\beta_k = k^{-4/7}$. At the $k$-th iteration, we compute the distance between the intermediate solution $\eta^k$ and $\eta^*$ by computing $\|\eta^k - \eta^*\|^2$, where $\eta^k$ is generated either by basic or accelerated SCGD. Then, we plot the $\log(k)$ versus $\log \|\eta^k - \eta^*\|$ for $k = 10,000$ to $40,000$ in Figure 2. The figure supports our theoretical analysis: a faster convergence of the accelerated SCGD is clearly seen.

## 5.2 Stochastic Shortest Path Problem

In this section, we consider a stochastic shortest path problem, as illustrated in Figure 3. Our objective is to minimize the expected length of path from node A to node B. At every node $i = 1, \ldots, n$, we have two possible actions: choose action $a_1$ and randomly move to an arrow-pointed adjacent node $j$ in Figure 3(i) with equal probability at a cost $c(i, j)$; choose action $a_2$ and randomly move to an arrow-pointed adjacent node $j$ in Figure 3(ii) with equal probability at a cost $\widehat{c}(i, j)$. The Bellman equation for this problem is

$$J(i) = \min \left\{ \mathbf{E}\big[c(i,j) + J(j) \mid a_1\big], \ \mathbf{E}\big[\widehat{c}(i,j) + J(j) \mid a_2\big] \right\}, \quad i = 1, \ldots, n, \quad J(B) = 0,$$

where the expectation is taken over possible future state $j$ conditioned on current state $i$ and the action $a_1$ or $a_2$. We denote by $J^* \in \Re^n$ the optimal solution to the Bellman equation, in which $J^*(i)$ is the optimal expected path length from node $i$ to node $B$. Solving the Bellman equation finds the optimal decision rule automatically.

Consider the Bellman residual minimization problem

$$\min_{J \in \Re^n} \sum_{i=1}^n \big(J(i) - \min\{q_{i,1}(J),\ q_{i,2}(J)\}\big)^2,$$

where we define $q_{i,1}, q_{i,2} \colon \Re^n \mapsto \Re$ to be functions given by

$$q_{i,1}(J) = \mathbf{E}\big[c(i,j) + J(j) \mid a_1\big], \qquad q_{i,2}(J) = \mathbf{E}\big[\widehat{c}(i,j) + J(j) \mid a_2\big].$$

We note that this problem is nonconvex. However, it is not hard to argue that the optimal $J^*$ is the only stationary solution. Therefore by applying the SCGD algorithm, we are guaranteed to find the optimal solution $J^*$. Note the "min" operation is nondifferentiable. In order to apply SCGD, we approximate $\max\{y_1, y_2\}$ by a differentiable function $h_\epsilon(y_1, y_2)$, as defined by

$$h_\epsilon(y_1, y_2) = \begin{cases} y_2 & \text{if } y_1 > y_2 + \epsilon, \\ y_1 & \text{if } y_2 > y_1 + \epsilon, \\ \frac{(y_1-y_2)^2}{4\epsilon} + \frac{y_1+y_2}{2} + \frac{\epsilon}{4} & \text{otherwise.} \end{cases}$$

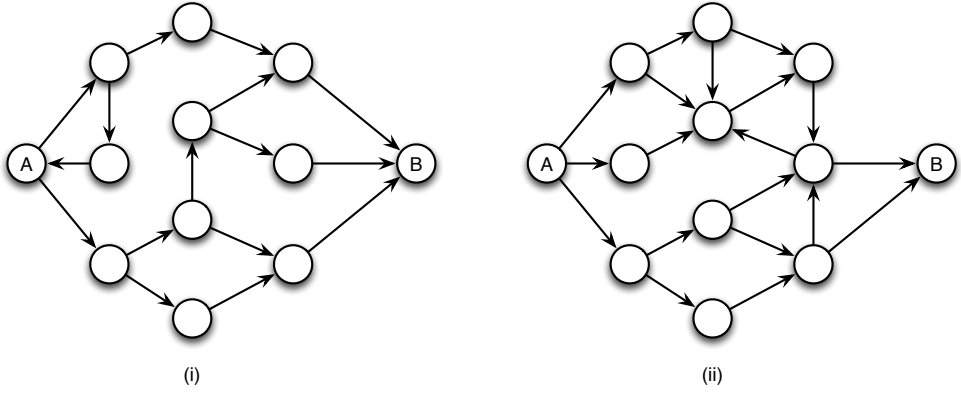

(i) (ii)

Figure 3: The stochastic shortest path problem. If we choose action $a_1$, we will randomly move to one of the adjacent nodes as in figure (i) with equal probability. If we choose action $a_2$, we will randomly move to one of the adjacent nodes as in figure (ii) with equal probability.

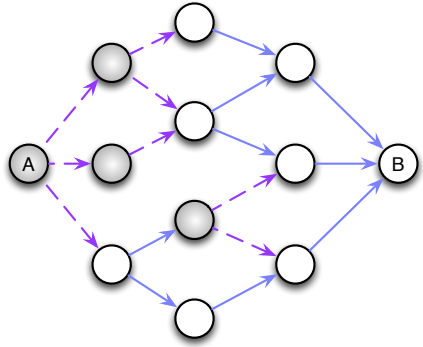

Figure 4: The optimal decision rule found by SCGD for the stochastic shortest path problem. If a node is colored white, the optimal action is $a_1$; if a node is colored grey, the optimal action is $a_2$. The transition probability matrix corresponding to the optimal decision rule is demonstrated by the arcs. The possible movements of action $a_1$ are blue solid arcs, and the possible movements of action $a_2$ are purple dashed arcs.

Let us consider the *smoothed* Bellman residual minimization, i.e.,

$$\min_{J \in \Re^n} \sum_{i=1}^{n} \Big(J(i) - h_\epsilon\big(q_{i,1}(J),\ q_{i,2}(J)\big)\Big)^2,$$

which takes the form of (1) if we take

$$f\big([J, Q_{1,1}, Q_{1,2} \ldots, Q_{n,1}, Q_{n,2}]\big) = \sum_{i=1}^{n} (J(i) - h_\epsilon(Q_{i,1}, Q_{i,2}))^2,$$

$$g(J) = [J, q_{1,1}(J), q_{1,2}(J) \ldots, q_{n,1}(J), q_{n,2}(J)],$$

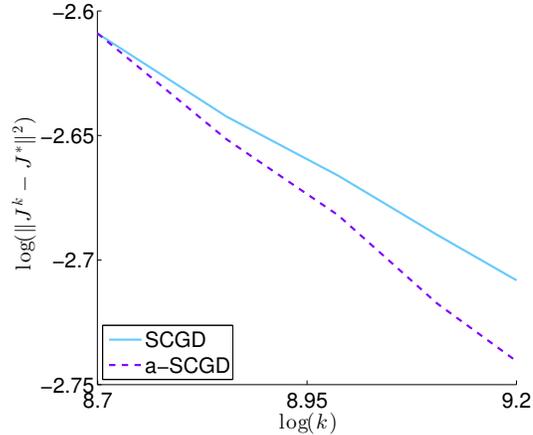

Figure 5: The empirical convergence rate of the estimate $J^k$ to the optimal $J^*$, where $J^k$ is computed by basic or accelerated SCGD.

where $g(J)$ is a mapping that involves expected values. Here $f$ is known and its gradient can be easily calculated. Given a simulator of the random walk, we can calculate sample gradients/values of $g$ by using simulated state transitions.

In the numerical experiment, we generate the weighted graph arbitrarily and build a simulator of the random walk. We apply the basic and accelerated SCGD algorithms using stepsizes suggested by preceding theorems. The SCGD algorithms update the estimate of $J^*$ every time after observing a new state transition. As a byproduct, the SCGD provides estimates of the $Q$ values as the algorithm proceeds. Given the optimal cost vector $J^*$ and associated $Q$ values, we easily obtain the optimal strategy by comparing wether $Q_{i,1} > Q_{i,2}$ for every node $i$. This yields an online decision strategy for the stochastic shortest path problem.

The optimal decision rule found by SCGD is demonstrated in Figure 4, which coincides with the optimal cost vector computed by the value iteration (see the textbook [4]). The convergence rate of the basic and accelerated SCGD algorithms is demonstrated in Figure 5. It plots $\log(\|J^k - J^*\|^2)$ against $\log(k)$, where $J^k$ is generated by the algorithm after observing $k$ samples. As shown in Figure 5, both algorithms converge to the optimal solution, and the accelerated SCGD indeed converges faster than the basic algorithm.

## 6 Conclusions

In this paper, we have considered the minimization of the composition of two expected-value functions. This problem involves two levels of stochasticity, so that the classical stochastic gradient no longer applies. We have proposed a class of stochastic compositional gradient methods that update based on random gradient evaluations of the inner and outer functions. We present a comprehensive convergence and rate of convergence analysis for these algorithms. A summary of the convergence rate and sample error complexity results is given in Table 1. For future research, one theoretical question is whether the current sample complexity is improvable or not. Another interesting direction is to extend the current work to optimization problems that involve compositions of more than two stochastic functions.